\documentclass{article}

\usepackage[final]{neurips_2018}

\usepackage[utf8]{inputenc} 
\usepackage[T1]{fontenc}    
\usepackage{hyperref}       
\usepackage{url}            
\usepackage{booktabs}       
\usepackage{nicefrac}       
\usepackage{microtype}      
\usepackage{natbib}
\usepackage{float}
\usepackage{amsmath, amsfonts, amssymb}
\usepackage{color}
\usepackage{graphicx}
\usepackage{makecell}
\usepackage{array, multirow}
\usepackage[symbol]{footmisc}

\renewcommand{\thefootnote}{\fnsymbol{footnote}}

\title{Critical initialisation for deep signal propagation in noisy rectifier neural networks}

\author{
  Arnu Pretorius\thanks{Correspondence: arnupretorius@gmail.com}  \\
  Computer Science Division \\
  CAIR\thanks{CSIR/SU Centre for Artificial Intelligence Research.} \\
  Stellenbosch University
  \And
  Elan Van Biljon \\
  Computer Science Division \\
  Stellenbosch University
  \And
  Steve Kroon \\
  Computer Science Division \\
  Stellenbosch University
  \And
  Herman Kamper \\
  Department of Electrical and Electronic Engineering \\
  Stellenbosch University\\
}

\usepackage[prependcaption,textsize=scriptsize]{todonotes}

\begin{document}

\maketitle

\begin{abstract}
    Stochastic regularisation is an important weapon in the arsenal of a deep learning practitioner.
    However, despite recent theoretical advances, our understanding of how noise influences signal propagation in deep neural networks remains limited.
    By extending recent work based on mean field theory, we develop a new framework for signal propagation in stochastic regularised neural networks.
    Our \textit{noisy signal propagation} theory can incorporate several common noise distributions, including additive and multiplicative Gaussian noise as well as dropout.
    We use this framework to investigate initialisation strategies for noisy ReLU networks.
    We show that no critical initialisation strategy exists using additive noise, with signal propagation exploding regardless of the selected noise distribution.
    For multiplicative noise (e.g.\ dropout), we identify alternative critical initialisation strategies that depend on the second moment of the noise distribution.
    Simulations and experiments on real-world data confirm that our proposed initialisation is able to stably propagate signals in deep networks, while using an initialisation disregarding noise fails to do so.
    Furthermore, we analyse correlation dynamics between inputs. Stronger noise regularisation is shown to reduce the depth to which discriminatory information about the inputs to a noisy ReLU network is able to propagate, even when initialised at criticality. We support our theoretical predictions for these trainable depths with simulations, as well as with experiments on MNIST and CIFAR-10.\footnote{Code to reproduce all the results is available at \href{https://github.com/ElanVB/noisy\_signal\_prop}{https://github.com/ElanVB/noisy\_signal\_prop}} 
    
\end{abstract}

\section{Introduction}

Over the last few years, advances in network design strategies have made it easier to train large networks and have helped to reduce overfitting.
These advances include improved weight initialisation strategies~\citep{glorot2010understanding, saxe2013exact, sussillo2014random, he2015delving, mishkin2015all}, non-saturating activation functions~\citep{glorot2011deep} and stochastic regularisation techniques~\citep{srivastava2014dropout}. 
Authors have noted, for instance, the critical dependence of successful training on noise-based methods such as dropout~\citep{krizhevsky2012imagenet,dahl2013improving}. 

In many cases, successful results arise only from effective combination of these advances.
Despite this interdependence, our theoretical understanding of how these mechanisms and their interactions affect neural networks remains impoverished. 




One approach to studying these effects is through the lens of deep neural signal propagation. 
By modelling the empirical input variance dynamics at the point of random initialisation, \citet{saxe2013exact} were able to derive equations capable of describing how signal propagates in nonlinear fully connected feed-forward neural networks. 
This ``mean field'' theory was subsequently extended by \citet{poole2016exponential} and \citet{schoenholz2016deep}, in particular, to analyse signal correlation dynamics. 
These analyses highlighted the existence of a critical boundary at initialisation, referred to as the ``edge of chaos''. This boundary defines a transition between ordered (vanishing), and chaotic (exploding) regimes for neural signal propagation. Subsequently, the mean field approximation to random neural networks has been employed to analyse other popular neural architectures \citep{yang2017mean, xiao2018dynamical, chen2018dynamical}. 

This paper focuses on the effect of noise on signal propagation in deep neural networks.
Firstly we ask: How is signal propagation in deep neural networks affected by noise?
To gain some insight into this question, we extend the mean field theory developed by \cite{schoenholz2016deep} for the special case of dropout noise, into a generalised framework capable of describing the signal propagation behaviour of stochastically regularised neural networks for different noise distributions.
 
Secondly we ask: How much are current weight initialisation strategies affected by noise-induced regularisation in terms of their ability to initialise at a critical point for stable signal propagation?
Using our derived theory, we investigate this question specifically for rectified linear unit (ReLU) networks.
In particular, we show that no such critical initialisation exists for arbitrary zero-mean additive noise distributions.
However, for multiplicative noise, such an initialisation is shown to be possible, given that it takes into account the amount of noise being injected into the network.
Using these insights, we derive novel critical initialisation strategies for several different multiplicative noise distributions.

Finally, we ask: Given that a network is initialised at criticality, in what way does noise influence the network's ability to propagate useful information about its inputs? 
By analysing the correlation between inputs as a function of depth in random deep ReLU networks, we highlight the following: even though the statistics of individual inputs are able to propagate arbitrarily deep at criticality, \textit{discriminatory information} about the inputs becomes lost at shallower depths as the noise in the network is increased. 
This is because in the later layers of a random noisy network, the internal representations from different inputs become uniformly correlated. Therefore, the application of noise regularisation directly limits the trainable depth of critically initialised ReLU networks.

\section{Noisy signal propagation}
\label{sec:noisysignalprop}

\renewcommand*{\thefootnote}{\arabic{footnote}}

 \begin{figure}
 \centering
 \includegraphics[width=100mm]{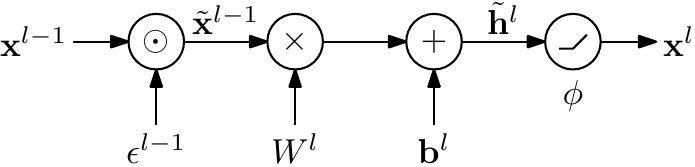}
 \caption{\textit{Noisy layer recursion.} 
The input $\mathbf{x}^{l-1}$ from the previous layer gets corrupted by the sampled noise $\epsilon^{l-1}$, either by vector addition or component-wise multiplication, producing the noisy inputs $\tilde{\mathbf{x}}^{l-1}$. 
The $l^{th}$ layer's corrupted pre-activations are then computed by multiplication with the layer weight matrix $W^l$, followed by a vector addition of the biases $\mathbf{b}^l$. 
Finally, the inputs to the next layer are simply the activations of the current layer, \textit{i.e.} $\mathbf{x}^l = \phi(\tilde{\mathbf{h}}^l)$. }
 \label{fig: noise diagram}
\end{figure}

We begin by presenting mean field equations for stochastically regularised fully connected feed-forward neural networks, allowing us to study noisy signal propagation for a variety of noise distributions. To understand how noise influences signal propagation in a random network given an input $\mathbf{x}^0 \in \mathbb{R}^{D_0}$, we inject noise into the model
\begin{align}
	\tilde{\mathbf{h}}^l = W^l(\mathbf{x}^{l-1}\odot \epsilon^{l-1}) + \mathbf{b}^l, \textcolor{white}{spa} \text{ for } l = 1, ..., L
	\label{eq: noisy deep model}
\end{align}
using the operator~$\odot$ to denote either addition or multiplication where $\epsilon^l$ is an input noise vector, sampled from a pre-specified noise distribution.
For additive noise, the distribution is assumed to be zero mean, for multiplicative noise distributions, the mean is assumed to be equal to one. 
The weights $W^l \in \mathbb{R}^{D_l \times D_{l-1}}$ and biases $\mathbf{b}^l \in \mathbb{R}^{D_l}$ are sampled i.i.d.\ from zero mean Gaussian distributions with variances $\sigma^2_w/D_{l-1}$ and $\sigma^2_b$, respectively, where $D_l$ denotes the dimensionality of the $l^{th}$ hidden layer in the network.
The hidden layer activations $\mathbf{x}^l = \phi(\tilde{\mathbf{h}}^l)$ are computed element-wise using an activation function $\phi(\cdot)$, 
for layers $l = 1, ..., L$.
Figure~\ref{fig: noise diagram} illustrates this recursive sequence of operations. 

To describe forward signal propagation for the model in \eqref{eq: noisy deep model}, we make use of the mean field approximation as in \cite{poole2016exponential}  and analyse the statistics of the internal representations of the network in expectation over the parameters and the noise. 
Since the weights and biases are sampled from zero mean Gaussian distributions with pre-specified variances, we can approximate the distribution of the pre-activations at layer $l$, in the large width limit, by a zero mean Gaussian with variance
\begin{align}
	\tilde{q}^l & = \sigma^2_w \left \{ \mathbb{E}_z \left [ \phi\left(\sqrt{\tilde{q}^{l-1}}z \right)^2 \right ] \odot \mu^{l-1}_2\right \} + \sigma^2_b, 
	\label{eq: noisy variance relation}
\end{align}
where $z \sim \mathcal{N}(0,1)$ (see Section \ref{sec: single input noisy signal prop} in the supplementary material). Here, $\mu^l_2 = \mathbb{E}_\epsilon[(\epsilon^{l})^2]$ is the second moment of the noise distribution being sampled from at layer $l$. The initial input variance is given by $q^0 = \frac{1}{D_0}\mathbf{x}^0\cdot\mathbf{x}^0$. 
Furthermore, to study the behaviour of a pair of signals from two different inputs, $\mathbf{x}^{0,a}$ and $\mathbf{x}^{0, b}$, passing through the network, we can compute the covariance at each layer as
\begin{align}
	\tilde{q}^l_{ab} = \sigma^2_w  \mathbb{E}_{z_1} \left [ \mathbb{E}_{z_2} \left [ \phi(\tilde{u}_1) \phi(\tilde{u}_2) \right ] \right ] + \sigma^2_b
	\label{eq: corr map}
\end{align}
where $\tilde{u}_1 = \sqrt{\tilde{q}^{l-1}_{aa}}z_1$ and $\tilde{u}_2 = \sqrt{\tilde{q}^{l-1}_{bb}}\left [ \tilde{c}^{l-1}z_1 + \sqrt{1-(\tilde{c}^{l-1})^2}z_2 \right ]$, with the correlation between inputs at layer $l$ given by $\tilde{c}^l = \tilde{q}^l_{ab}/\sqrt{\tilde{q}^l_{aa}\tilde{q}^l_{bb}}$. Here, $q^l_{aa}$ is the variance of $\tilde{\mathbf{h}}^{l,a}_j$ (see Section \ref{sec: two input noisy signal prop} in the supplementary material for more details).

For the backward pass, we use the equations derived in \cite{schoenholz2016deep} to describe error signal propagation.\footnote[1]{It is, however, important to note that the derivation relies on the assumption that the weights used in the forward pass are sampled independently from those used during backpropagation.} In the context of mean field theory, the expected magnitude of the gradient at each layer can be shown to be proportional to the variance of the error, $\tilde{\delta}^l_i = \phi^\prime(\tilde{\mathbf{h}}^l_i)\sum^{D_{l+1}}_{j=1}\tilde{\delta}^{l+1}_jW^{l+1}_{ji}$. This allows for the distribution of the error signal at layer $l$ to be approximated by a zero mean Gaussian with variance 
\begin{align}
	\tilde{q}^l_\delta = \tilde{q}^{l+1}_\delta\frac{D_{l+1}}{D_l}\sigma^2_w \mathbb{E}_z \left [ \phi^{\prime}\left(\sqrt{\tilde{q}^l}z\right)^2 \right ].
	\label{eq: error variance relation}
\end{align}
Similarly, for noise regularised networks, the covariance between error signals can be shown to be
\begin{align}
	\tilde{q}^l_{ab, \delta} = \tilde{q}^{l+1}_{ab, \delta} \frac{D_{l+1}}{D_{l+2}}\sigma^2_w \mathbb{E}_{z_1} \left [ \mathbb{E}_{z_2} \left [ \phi^{\prime}(\tilde{u}_1) \phi^{\prime}(\tilde{u}_2) \right ] \right ],
	\label{eq: error covariance relation}
\end{align}
where $\tilde{u}_1$ and $\tilde{u}_2$ are defined as was done in the forward pass. 

\begin{figure}
	\centering
	\includegraphics[width=0.9\linewidth]{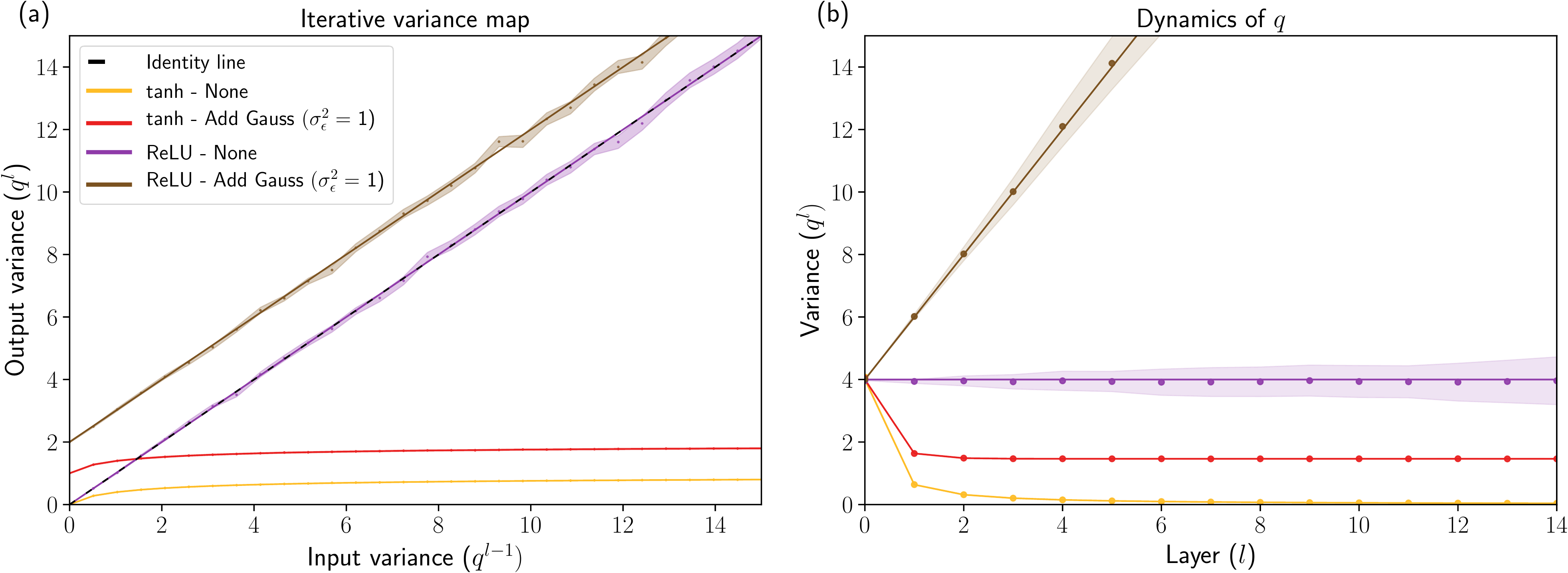}
	\caption{\textit{Deep signal propagation with and without noise}. \textbf{(a)}: Iterative variance map. \textbf{(b)}: Variance dynamics during forward signal propagation. In (a) and (b), lines correspond to theoretical predictions
	and points to numerical simulations (means over $50$ runs with shaded one standard deviation bounds), for noiseless tanh (yellow) and noiseless ReLU (purple) networks,
	as well as for noisy tanh (red) and noisy ReLU (brown) networks regularised using additive noise from a standard Gaussian.
	Both tanh networks use $(\sigma_w , \sigma_b ) = (1, 0)$, the ``Xavier'' initialisation \citep{glorot2010understanding}, while the ReLU networks use $(\sigma_w, \sigma_b) = (\sqrt{2}, 0)$ the ``He'' initialisation \citep{he2015delving}. In our experiments, we use network layers consisting of $1000$ hidden units (see  Section \ref{sec: experimental details} in the supplementary material for more details on all our simulated experiments). }
	\label{fig: add noise qmap}
\end{figure}

Equations \eqref{eq: noisy variance relation}-\eqref{eq: error covariance relation} fully capture the relevant statistics that govern signal propagation for a random network during both the forward and the backward pass.
In the remainder of this paper, we consider, as was done by \cite{schoenholz2016deep}, the following necessary condition for training: ``for a random network to be trained information about the inputs should be able to propagate
forward through the network, and information about the gradients should be able to propagate
backwards through the network.''
The behaviour of the network at this stage depends on the choice of activation, noise regulariser and initial parameters. 
In the following section, we will focus on networks that use the Rectified Linear Unit (ReLU) as activation function. 
The chosen noise regulariser is considered a design choice left to the practitioner. 
Therefore, whether a random noisy ReLU network satisfies the above stated necessary condition for training largely depends on the starting parameter values of the network, \textit{i.e.} its initialisation.





\section{Critical initialisation for noisy rectifier networks}

Unlike the tanh nonlinearity investigated in previous work \citep{poole2016exponential,schoenholz2016deep}, rectifying activation functions such as ReLU are unbounded.
This means that the statistics of signal propagation through the network is not guaranteed to naturally stabilise through saturating activations, as shown in Figure \ref{fig: add noise qmap}.

A point on the identity line in Figure \ref{fig: add noise qmap}~(a) represents a fixed point to the recursive variance map in equation \eqref{eq: noisy variance relation}. At a fixed point, signal will stably propagate through the remaining layers of the network. 
For tanh networks, such a fixed point always exists irrespective of the initialisation, or the amount of noise injected into the network.
For ReLU networks, this is not the case.
Consider the ``He'' initialisation \citep{he2015delving} for ReLU, commonly used in practice. 
In (b), we plot the variance dynamics for this initialisation in purple and observe stable behaviour. 
But what happens when we inject noise into each network?
In the case of tanh (shown in red), the added noise simply shifts the fixed point to a new stable value. However, for ReLU, the noise entirely destroys the fixed point for the ``He'' initialisation, making signal propagation unstable. 
This can be seen in (a), where the variance map for noisy ReLU (shown in brown) moves off the identity line entirely, causing the signal in (b) to explode.

Therefore, to investigate whether signal can stably propagate through a random \textit{noisy} ReLU network, we examine \eqref{eq: noisy variance relation} more closely, which for ReLU becomes (see Section \ref{sec: var dynamics for noisy ReLU networks} in supplementary material)
\begin{align}
	\tilde{q}^l = \sigma^2_w \left [\frac{\tilde{q}^{l-1}}{2} \odot \mu_2 \right ]  + \sigma^2_b.
	\label{eq: relu variance map}                                                
\end{align}
For ease of exposition we assume equal noise levels at each layer, \textit{i.e.} $\mu^{l}_2 = \mu_2, \forall l$. A critical initialisation for a noisy ReLU network occurs when the tuple $(\sigma_w, \sigma_b, \mu_2)$ provides a fixed point $\tilde{q}^*$, to the recurrence in \eqref{eq: relu variance map}. 
This at least ensures that the statistics of individual inputs to the network will be preserved throughout the first forward pass.
The existence of such a solution depends on the type of noise that is injected into the network.
In the case of additive noise, $\tilde{q}^* = \sigma^2_w\frac{1}{2}\tilde{q}^* + \mu_2 \sigma^2_w + \sigma^2_b$, implying that the only critical point initialisation for non-zero $\tilde{q}^*$ is given by $(\sigma_w, \sigma_b, \mu_2) = (\sqrt{2}, 0, 0)$.
Therefore, critical initialisation is not possible using any amount of zero-mean additive noise, regardless of the noise distribution.
For multiplicative noise, $\tilde{q}^* = \sigma^2_w\frac{1}{2}\tilde{q}^*\mu_2 + \sigma^2_b$, so the solution $(\sigma_w, \sigma_b, \mu_2) = \left ( \sqrt{\frac{2}{\mu_2}}, 0, \mu_2 \right)$ provides a critical initialisation for noise distributions with mean one and a non-zero second moment~$\mu_2$. 
For example, in the case of multiplicative Gaussian noise, $\mu_2 = \sigma^2_\epsilon + 1$, yielding critical initialisation with $(\sigma_w, \sigma_b) = \left (\sqrt{\frac{2}{\sigma^2 + 1}}, 0 \right)$.
For dropout noise, $\mu_2 = 1/p$ (with $p$ the probability of retaining a neuron); thus, to initialise at criticality, we must set $(\sigma_w, \sigma_b) = (\sqrt{2p}, 0)$.
Table \ref{tab: crit init table} summarises critical initialisations for some commonly used noise distributions. 
We also note that similar results can be derived for other rectifying activation functions; for example, for multiplicative noise the critical initialisation for parametric ReLU (PReLU) activations (with slope parameter $\alpha$) is given by $(\sigma_w, \sigma_b, \mu_2) = \left ( \sqrt{\frac{2}{\mu_2(\alpha^2 + 1)}}, 0, \mu_2 \right)$.

\begin{table}[t]
	\caption{Critical point initialisation for noisy ReLU networks.}
	\label{tab: crit init table}
	\vskip 0.15in
	\begin{center}
		\begin{small}
			\begin{sc}
				\begin{tabular}{llll}
					\toprule
					Distribution & $\textrm{p}(\epsilon)$ &  $\mu_2$  & critical initialisation                                                                                      \\
					\toprule
					--- Additive Noise ---  \\
					\midrule
					Gaussian   & $\mathcal{N}(0, \sigma^2_\epsilon) $  & $\sigma^2_\epsilon$ \vspace{1em} & $(\sigma_w, \sigma_b, \sigma_\epsilon) = (\sqrt{2}, 0, 0)$                                                         \\
					Laplace   & $Lap(0, \beta)$   &   $2\beta^2$ \vspace{0.3em}  & $(\sigma_w, \sigma_b, \beta) = (\sqrt{2}, 0, 0)$                                                                   \\
					\toprule
					--- Multiplicative Noise ---  \\
					\midrule
					Gaussian   & $\mathcal{N}(1, \sigma^2_\epsilon) $  &   $(\sigma^2_\epsilon + 1)$ \vspace{1em} & $(\sigma_w, \sigma_b, \sigma_\epsilon) = \left ( \sqrt{\frac{2}{\sigma^2_\epsilon+1}}, 0, \sigma_\epsilon \right)$ \\
					Laplace  & $Lap(1, \beta)$    &  $(2\beta^2 + 1)$ \vspace{1em}  & $(\sigma_w, \sigma_b, \beta) = \left(\sqrt{\frac{2}{2\beta^2 + 1}}, 0, \beta \right)$                              \\
					Poisson   & $Poi(1)$   &   $2$ \vspace{1em}  & $(\sigma_w, \sigma_b, \lambda) = \left(1, 0, 1 \right)$                                                            \\
					Dropout    & \makecell[l]{$\textrm{p}(\epsilon = \frac{1}{p}) = p,$\\$\textrm{p}(\epsilon = 0) = 1-p$ }  &   $\frac{1}{p}$ \vspace{0.3em} & $(\sigma_w, \sigma_b, p) = (\sqrt{2p}, 0, p)$                                                                      \\
					\bottomrule
				\end{tabular}
			\end{sc}
		\end{small}
	\end{center}
	\vskip -0.1in
\end{table}

\begin{figure}
	\includegraphics[width=\linewidth]{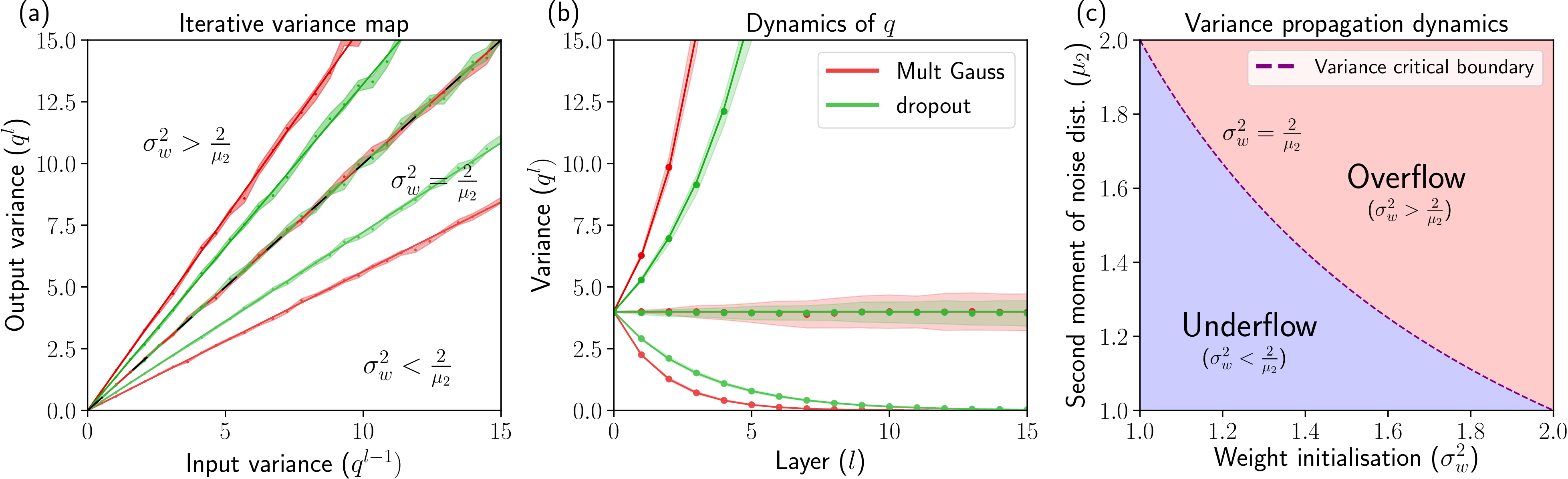}
	\caption{\textit{Critical initialisation for noisy ReLU networks.} \textbf{(a)}: Iterative variance map. \textbf{(b)}: Variance dynamics during forward signal propagation. In (a) and (b), lines correspond to theoretical predictions and points to numerical simulations. Dropout ($p=0.6$) is shown in green for different initialisations, $\sigma^2_w = 2(0.6) = \frac{2}{\mu_2}$ (critical), $\sigma^2_w = (1.15)^2 \frac{2}{(0.6)^{-1}} > \frac{2}{\mu_2}$ (exploding signal) and $\sigma^2_w = (0.85)^2 \frac{2}{(0.6)^{-1}} < \frac{2}{\mu_2}$ (vanishing signal). Similarly, multiplicative Gaussian noise ($\sigma_\epsilon = 0.25$) is shown in red with $\sigma^2_w = \frac{2}{(0.25)^2 + 1} = \frac{2}{\mu_2}$ (critical), $\sigma^2_w = (1.25)^2 \frac{2}{
		\mu_2
	}$ (exploding) and $\sigma^2_w = (0.75)^2 \frac{2}{\mu_2}$ ( vanishing). \textbf{(c)}: Variance critical boundary for initialisation, separating numerical overflow and underflow signal propagation regimes.}
	\label{fig: qmap}
\end{figure}

To see the effect of initialising on or off the critical point for ReLU networks, Figure \ref{fig: qmap} compares the predicted versus simulated variance dynamics for different initialisation schemes. 
For schemes not initialising at criticality, the variance map in (a) no longer lies on the identity line and as a result the forward propagating signal in (b) either explodes, or vanishes. 
In contrast, the initialisations derived above lie on the critical boundary between these two extremes, as shown in (c) as a function of the noise. 
By compensating for the amount of injected noise, the signal corresponding to the initialisation $\sigma^2_w = \frac{2}{\mu_2}$ is preserved in (b) throughout the entire forward pass, with roughly constant variance dynamics.

\begin{figure}
	\includegraphics[width=\linewidth]{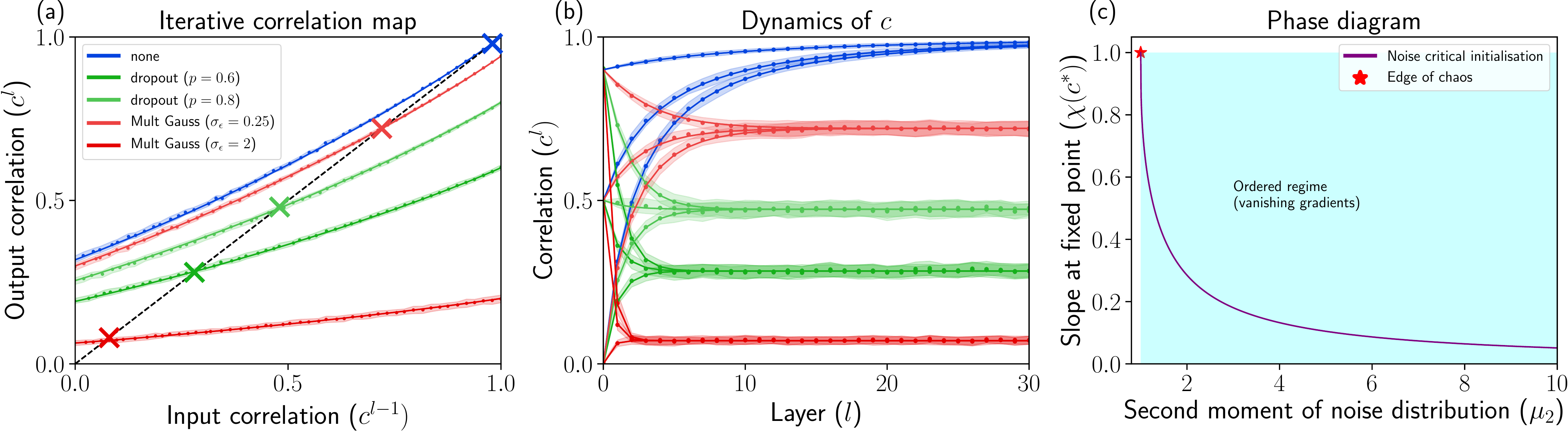}
	\caption{\textit{Propagating correlation information in noisy ReLU networks.} \textbf{(a)}: Iterative correlation map with fixed points indicated by ``X'' marks on the identity line. \textbf{(b)}: Correlation dynamics during forward signal propagation. In (a) and (b), lines correspond to theoretical predictions and points to numerical simulations. All simulated networks were initialised at criticality for each noise type and level. \textbf{(c)}: Slope at the fixed point correlation as a function of the amount of noise injected into the network.}
	\label{fig: cmap}
\end{figure}

Next, we investigate the correlation dynamics between inputs. 
Assuming that \eqref{eq: relu variance map} is at its fixed point $\tilde{q}^*$, which exists only if $\sigma^2_w = \frac{2}{\mu_2}$, the correlation map for a noisy ReLU network is given by (see Section \ref{sec: correlation dynamics for noisy ReLU networks} in supplementary material)
\begin{align}
	\tilde{c}^l = \frac{1}{\mu_2} \left \{ \frac{\tilde{c}^{l-1}\text{sin}^{-1} \left ( \tilde{c}^{l-1} \right ) + \sqrt{1-(\tilde{c}^{l-1})^2}}{\pi} 
	+ \frac{\tilde{c}^{l-1}}{2} \right \}.                                                                                                            
\end{align} 
Figure \ref{fig: cmap} plots this theoretical correlation map against simulated dynamics for different noise types and levels. 
For no noise, the fixed point $c^*$ in (a) is situated at one (marked with an ``X'' on the blue line). 
The slope of the blue line indicates a non-decreasing function of the input correlations. 
After a certain depth, inputs end up perfectly correlated irrespective of their starting correlation, as shown in (b). 
In other words, random deep ReLU networks lose discriminatory information about their inputs as the depth of the network increases, even when initialised at criticality. 
When noise is added to the network, inputs decorrelate and $c^*$ moves away from one. 
However, more importantly, correlation information in the inputs become lost at shallower depths as the noise level increases, as can be seen in (b). 

\begin{figure}
	\centering
	\includegraphics[width=0.9\linewidth]{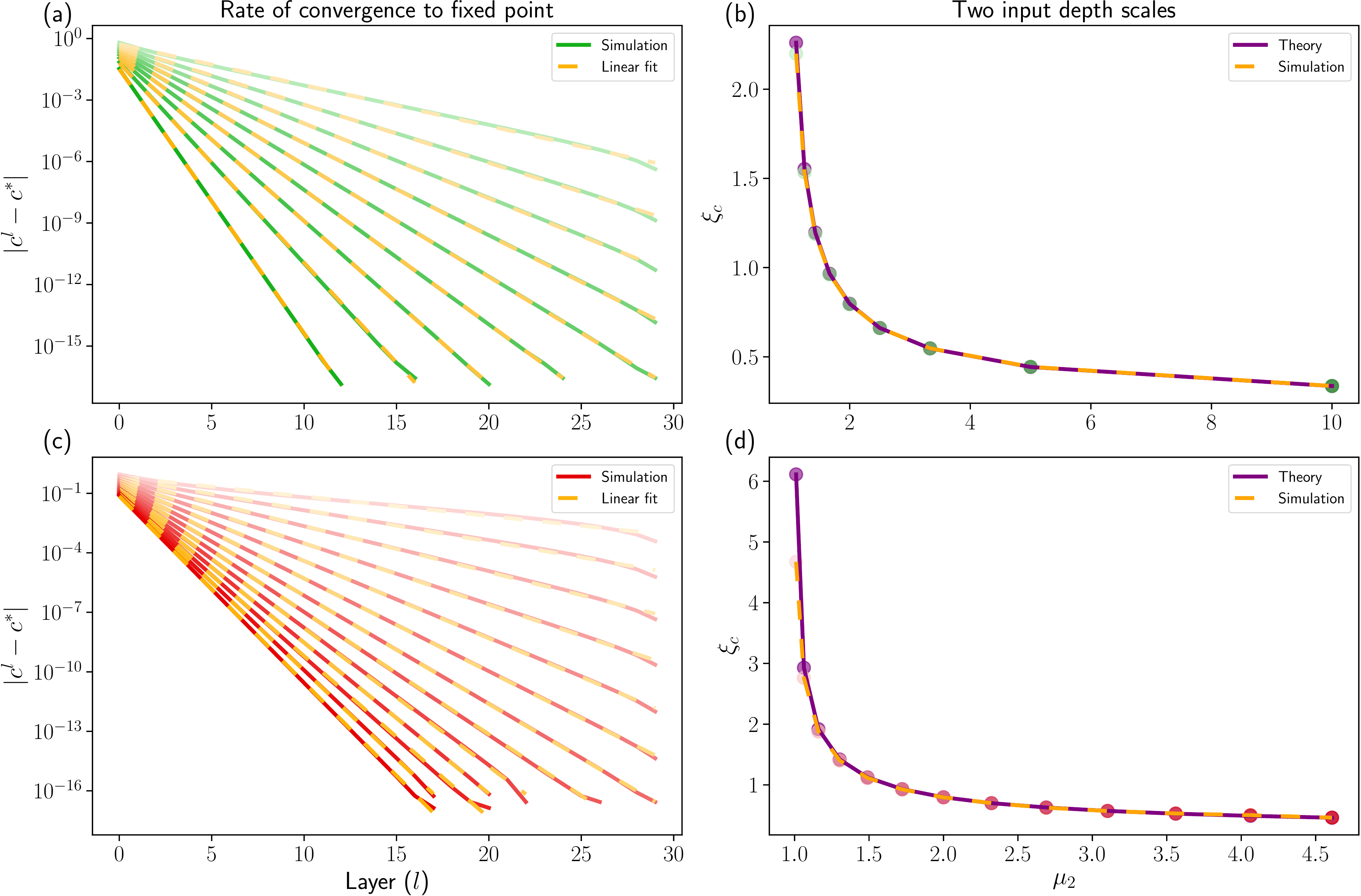}
	\caption{\textit{Noise dependent depth scales for training.} \textbf{(a)}: Linear fits (dashed lines) to $|c^l - c^*|$ as a function of depth on a log-scale (solid lines) for varying amounts of dropout ($p=0.1$ to $p=0.9$ by $0.1$). \textbf{(b)}: Theoretical depth scales (solid lines) versus empirically inferred scales (dashed lines) per dropout rate. Scales are inferred noting that if $|c^l - c^*| \sim e^{-l/\xi_c}$, then a linear fit, $al + b$, in the logarithmic domain gives $\xi_c \approx -\frac{1}{a}$, for large $l$. In other words, the negative inverse slope of a linear fit to the log differences in
	correlation should match the theoretical values for $\xi_c$. Therefore, we compare $\xi_c = -1/\text{ln}\left [ \chi(c^*) \right ]$ to $-\frac{1}{a}$ for different levels of noise. \textbf{(c) - (d)}: Similar to (a) and (b), but for Gaussian noise ($\sigma_\epsilon = 0.1$ to $\sigma_\epsilon = 1.9$ by $0.15$).}
	\label{fig: theory depth scales}
\end{figure}

How quickly a random network loses information about its inputs depends on the rate of convergence to the fixed point $c^*$. 
Using this observation, \cite{schoenholz2016deep} derived so-called depth scales $\xi_c$, by assuming $|c^l - c^*| \sim e^{-l/\xi_c}$. 
These scales essentially control the feasible depth at which networks can be considered trainable, since they may still allow useful correlation information to propagate through the network.
In our case, the depth scale for a noisy ReLU network under this assumption can be shown to be (see Section \ref{sec: Limiting trainability depth scales} in supplementary material)
\begin{align}
	\xi_c = -1/\text{ln}\left [ \chi(c^*) \right ], 
	\label{eq: depth scale}
\end{align}
where
\begin{align}
	\chi(c^*) = \frac{1}{\mu_2\pi} \left [ \text{sin}^{-1}\left( c^* \right) + \frac{\pi}{2} \right ]. 
\end{align}
The exponential rate assumption underlying the derivation of \eqref{eq: depth scale} is supported in Figure \ref{fig: theory depth scales}, where for different noise types and levels, we plot $|c^l - c^*|$ as a function of depth on a log-scale, with corresponding linear fits (see panels (a) and (c)). 
We then compare the theoretical depth scales from \eqref{eq: depth scale} to actual depth scales obtained through simulation (panels (b) and (d)), as a function of noise and observe a good fit for non-zero noise levels.\footnote{We note \cite{hayou2018selection} recently showed that the rate of convergence for noiseless ReLU networks is not exponential, but polynomial instead. Interestingly, keeping with the exponential rate assumption, we indeed find that the discrepancy between our theoretical depth scales from \eqref{eq: depth scale} and our simulated depth scales, is largest at very low noise levels. However, at more typical noise levels, such as a dropout rate of $p=0.5$ for example, the assumption seems to provide a close fit, with good agreement between theory and simulation.}
We thus find that noise limits the depth at which critically initialised ReLU networks are expected to perform well through training.



We next briefly discuss error signal propagation during the backward pass for noise regularised ReLU networks. When critically initialised, the error variance recurrence relation in \eqref{eq: error variance relation} for these networks is (see Section \ref{sec: backpropagation dynamics for a single input} in supplementary material)
\begin{align}
	\tilde{q}^l_\delta = \tilde{q}^{l+1}_\delta\frac{D_{l+1}}{D_l\mu_2}, 
	\label{eq: error variance relation relu}                             
\end{align}
with the covariance between error signals in \eqref{eq: error covariance relation}, given by (see Section \ref{sec: backpropagation dynamics for a pair of inputs} in supplementary material)
\begin{align}
	\tilde{q}^l_{ab, \delta} = \tilde{q}^{l+1}_{ab, \delta} \frac{D_{l+1}}{D_{l+2}}\chi(c^*). 
	\label{eq: error correlation relation relu}                                               
\end{align}
Note the explicit dependence on the width of the layers of the network in \eqref{eq: error variance relation relu} and \eqref{eq: error correlation relation relu}. We first consider constant width networks, where $D_{l+1} = D_l$, for all $l = 1, ..., L$. 
For any amount of multiplicative noise, $\mu_2 > 1$, and we see from \eqref{eq: error variance relation relu} that gradients will tend to vanish for large depths. 
Furthermore, Figure \ref{fig: cmap} (c) plots $\chi(c^*)$ as a function of $\mu_2$.
As $\mu_2$ increases from one, $\chi(c^*)$ decreases from one.
Therefore, from \eqref{eq: error correlation relation relu}, we also find that error signals from different inputs will tend to decorrelate at large depths.

Interestingly, for non-constant width networks, stable gradient information propagation may still be possible. 
If the network architecture adapts to the amount of noise being injected by having the widths of the layers grow as $D_{l+1} = D_l\mu_2$, then \eqref{eq: error variance relation relu} should be at its fixed point solution. 
For example, in the case of dropout $D_{l+1} = D_l/p$, which implies that for any $p < 1$, each successive layer in the network needs to grow in width by a factor of $1/p$ to promote stable gradient flow. Similarly, for multiplicative Gaussian noise, $D_{l+1} = D_l(\sigma^2_\epsilon + 1)$, which requires the network to grow in width unless $\sigma^2_\epsilon = 0$. 
Similarly, if $D_{l+2} = D_{l+1}\chi(c^*) = D_l \mu_2 \chi(c^*)$ in \eqref{eq: error correlation relation relu}, the covariance of the error signal should be preserved during the backward pass, for arbitrary values of $\mu_2$ and $\chi(c^*)$.

\section{Experimental results}

From our analysis of deep noisy ReLU networks in the previous section, we expect that a necessary condition for such a network to be trainable, is that the network be initialised at criticality. However, whether the layer widths are varied or not for the sake of backpropagation, the correlation dynamics in the forward pass may still limit the depth at which these networks perform well. 

\begin{figure}
	\includegraphics[width=\linewidth]{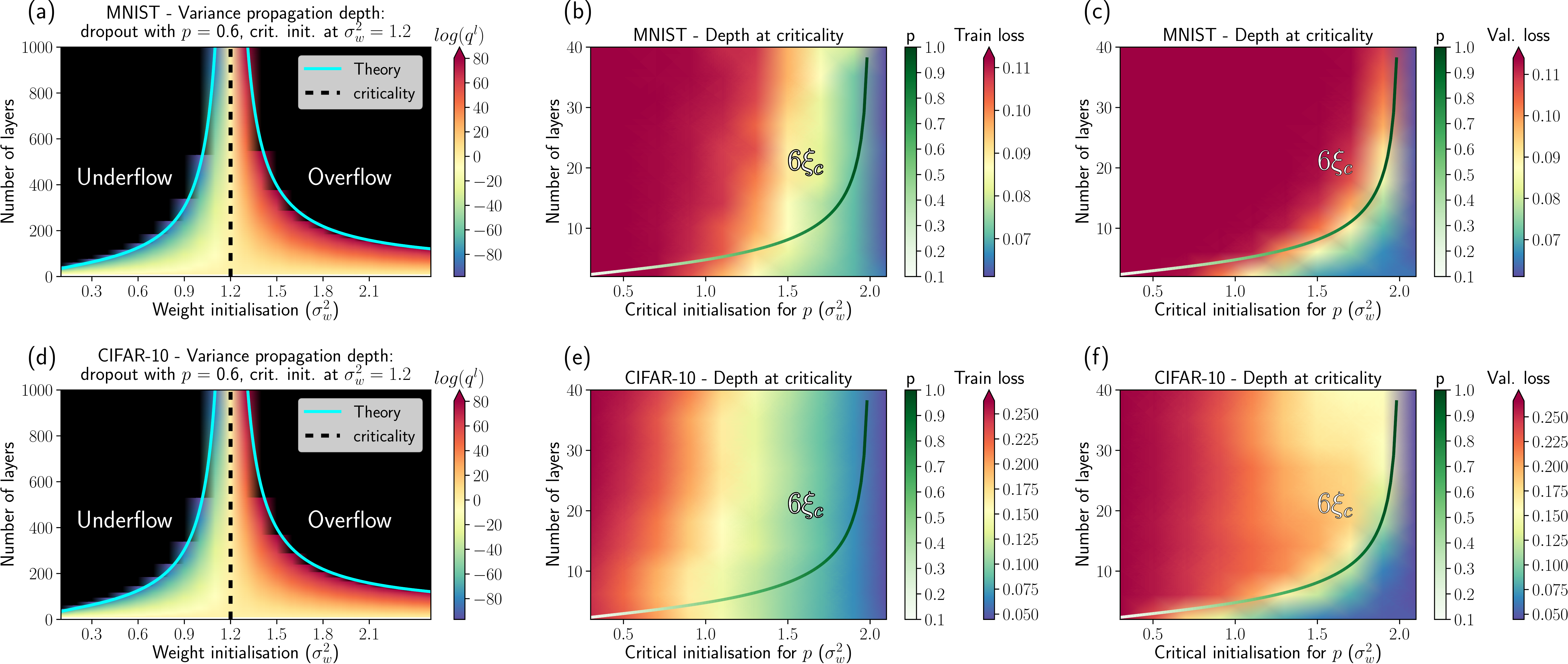}
	\caption{\textit{Depth scale experiments on MNIST and CIFAR-10.} \textbf{(a)} Variance propagation dynamics for MNIST on and off the critical point initialisation (dashed black line) with dropout ($p = 0.6$). The cyan curve represents the theoretical boundary at which numerical instability issues are predicted to occur and is computed as $L^* = \text{ln}(K)/\text{ln}(\frac{\sigma^2_w}{2}\mu_2)$, where $K$ is the largest (or smallest) positive number representable by the computer. Specifically, we use $32$-bit floating point numbers and set $K = 3.4028235\times 10^{38}$, if $\sigma^2_w > \frac{2}{\mu_2}$ and $K = 1.1754944\times 10^{-38}$, if $\sigma^2_w < \frac{2}{\mu_2}$. \textbf{(b)} Depth scales fit to the training loss on MNIST for networks initialised at criticality for dropout rates $p=0.1$ (severe dropout) to $p=1$ (no dropout). \textbf{(c)} Depth scales fit to the validation loss on MNIST. \textbf{(d) - (f)}: Similar to (a) - (c), but for CIFAR-10. For each plot we highlight trends by smoothing the colour grid (for non smoothed versions see Section \ref{sec: additional results} in the supplementary material).} 
	\label{fig: depth scales}
\end{figure}

We therefore investigate the performance of noise-regularised deep ReLU networks on real-world data. 
First, we validate the derived critical initialisation.
As the depth of the network increases, any initialisation strategy that does not factor in the effects of noise, will cause the forward propagating signal to  become increasingly unstable. 
For very deep networks, this might cause the signal to either explode or vanish, even within the first forward pass, making the network untrainable. 
To test this, we sent inputs from MNIST and CIFAR-10 through ReLU networks using dropout (with $p = 0.6$) at varying depths and for different initialisations of the network. 
Figure \ref{fig: depth scales} (a) and (d) shows the evolution of the input statistics as the input propagates through each network for the different data sets. 
For initialisations not at criticality, the variance grows or shrinks rapidly to the point of causing numerical overflow or underflow (indicated by black regions). For deep networks, this can happen well before any signal is able to reach the output layer. 
In contrast, initialising at criticality (as shown by the dashed black line), allows for the signal to propagate reliably even at very large depths.
Furthermore, given the floating point precision, if $\sigma^2_w \neq \frac{2}{\mu_2}$, we can predict the depth at which numerical overflow (or underflow) will occur by solving for $L^*$ in $K = \left(\sigma^2_w\mu_2/2 \right )^{L^*}q^0$, where $K$ is the largest (or smallest) positive number representable by the computer (see Section \ref{sec: var depth scale theory} in supplementary material). These predictions are shown by the cyan line and provide a good fit to the empirical limiting depth from numerical instability.

We now turn to the issue of limited trainability. 
Due to the loss of correlation information between inputs as a function of noise and network depth, we expect noisy ReLU networks not to be able to perform well beyond certain depths. 
We investigated depth scales for ReLU networks with dropout initialised at criticality: we trained $100$ networks on MNIST and CIFAR-10 for $200$ epochs using SGD and a learning rate of $10^{-3}$ with dropout rates ranging from $0.1$ to $1$ for varying depths. 
The results are shown in Figure \ref{fig: depth scales} (see Section \ref{sec: additional results} of the supplementary material for additional experimental results). 
For each network configuration and noise level, the critical initialisation $\sigma^2_w = \frac{2}{\mu_2}$ was used. 
We indeed observe a relationship between depth and noise on the loss of a network, even at criticality. Interestingly, the line $6\xi_c$ \citep{schoenholz2016deep}, seems to track the depth beyond which the relative performance on the validation loss becomes poor, more so than on the training loss. However, in both cases, we find that even modest amounts of noise can limit performance.

\section{Discussion}
\label{sec:discussion}

By developing a general framework to study signal propagation in noisy neural networks, we were able to show how different stochastic regularisation strategies may impact the flow of information in a deep network. 
Focusing specifically on ReLU networks, we derived novel critical initialisation strategies for multiplicative noise distributions and showed that no such critical initialisations exist for commonly used additive noise distributions. 
At criticality however, our theory predicts that the statistics of the input should remain within a stable range during the forward pass and enable reliable signal propagation for noise regularised deep ReLU networks. 
We verified these predictions by comparing them with numerical simulations as well as experiments on MNIST and CIFAR-10 using dropout and found good agreement. 

Interestingly, we note that a dropout rate of $p = 0.5$ has often been found to work well for ReLU networks \citep{srivastava2014dropout}. 
The critical initialisation corresponding to this rate is $(\sigma_w, \sigma_b) = (\sqrt{2p}, 0) = (1, 0)$. This is exactly the ``Xavier'' initialisation proposed by \citet{glorot2010understanding}, which prior to the development of the ``He'' initialisation, was often used in combination with dropout~\citep{simonyan2014very}. 
This could therefore help to explain the initial success associated with this specific dropout rate. Similarly, \citet{srivastava2014dropout} reported that adding multiplicative Gaussian noise where $\epsilon \sim \mathcal{N}(1, \sigma^2_\epsilon)$, with $\sigma^2_\epsilon = 1$, also seemed to perform well, for which the critical initialisation is $\left ( \sqrt{\frac{2}{\sigma^2_\epsilon+1}}, 0 \right) = (1, 0)$, again corresponding to the ``Xavier'' method.

Although our initialisations ensure that individual input statistics are preserved, we further analysed the correlation dynamics between inputs and found the following: at large depths inputs become predictably correlated with each other based on the amount of noise injected into the network. 
As a consequence, the representations for different inputs to a deep network may become indistinguishable from each other in the later layers of the network. This can make training infeasible for noisy ReLU networks of a certain depth and depends on the amount of noise regularisation being applied.

We now note the following shortcomings of our work: firstly, our findings only apply to fully connected feed-forward neural networks and focus almost exclusively on the ReLU activation function. 
Furthermore, we limit the scope of our architectural design to a recursive application of a dense layer followed by a noise layer, whereas in practice a larger mix of layers is usually required to solve a specific task. 

Ultimately, we are interested in reducing the number of decisions that need to made when designing deep neural networks and understanding the implications of those decisions on network behaviour and performance. 
Any machine learning engineer exploring a neural network based solution to a practical problem will be faced with a large number of possible design decisions. 
All these decisions cost valuable time to explore. 
In this work, we hope to have at least provided some guidance in this regard, specifically when choosing between different initialisation strategies for noise regularised ReLU networks and understanding their associated implications.

\section*{Acknowledgements}



We would like to thank the reviewers for their insightful comments which improved the quality of this work. 
Furthermore, we would like to thank Google, the CSIR/SU Centre for Artificial Intelligence Research (CAIR) as well as the Science Faculty and the Postgraduate and International Office of Stellenbosch University for financial support. Finally, we gratefully acknowledge the support of NVIDIA Corporation with the donation of a Titan Xp GPU used for this research.

\bibliographystyle{IEEEtranN}
\bibliography{bibfile}

\begin{thebibliography}{16}
\providecommand{\natexlab}[1]{#1}
\providecommand{\url}[1]{#1}
\csname url@samestyle\endcsname
\providecommand{\newblock}{\relax}
\providecommand{\bibinfo}[2]{#2}
\providecommand{\BIBentrySTDinterwordspacing}{\spaceskip=0pt\relax}
\providecommand{\BIBentryALTinterwordstretchfactor}{4}
\providecommand{\BIBentryALTinterwordspacing}{\spaceskip=\fontdimen2\font plus
\BIBentryALTinterwordstretchfactor\fontdimen3\font minus
  \fontdimen4\font\relax}
\providecommand{\BIBforeignlanguage}[2]{{%
\expandafter\ifx\csname l@#1\endcsname\relax
\typeout{** WARNING: IEEEtranN.bst: No hyphenation pattern has been}%
\typeout{** loaded for the language `#1'. Using the pattern for}%
\typeout{** the default language instead.}%
\else
\language=\csname l@#1\endcsname
\fi
#2}}
\providecommand{\BIBdecl}{\relax}
\BIBdecl

\bibitem[Glorot and Bengio(2010)]{glorot2010understanding}
X.~Glorot and Y.~Bengio, ``Understanding the difficulty of training deep
  feedforward neural networks,'' in \emph{Proceedings of the International
  Conference on Artificial Intelligence and Statistics}, 2010, pp. 249--256.

\bibitem[Saxe et~al.(2014)Saxe, McClelland, and Ganguli]{saxe2013exact}
A.~M. Saxe, J.~L. McClelland, and S.~Ganguli, ``Exact solutions to the
  nonlinear dynamics of learning in deep linear neural networks,''
  \emph{Proceedings of the International Conference on Learning
  Representations}, 2014.

\bibitem[Sussillo and Abbott(2014)]{sussillo2014random}
D.~Sussillo and L.~Abbott, ``Random walk initialization for training very deep
  feedforward networks,'' \emph{arXiv preprint arXiv:1412.6558}, 2014.

\bibitem[He et~al.(2015)He, Zhang, Ren, and Sun]{he2015delving}
K.~He, X.~Zhang, S.~Ren, and J.~Sun, ``Delving deep into rectifiers: Surpassing
  human-level performance on {ImageNet} classification,'' in \emph{Proceedings
  of the IEEE International Conference on Computer Vision}, 2015, pp.
  1026--1034.

\bibitem[Mishkin and Matas(2016)]{mishkin2015all}
D.~Mishkin and J.~Matas, ``All you need is a good init,'' \emph{Proceedings of
  International Conference on Learning Representations}, 2016.

\bibitem[Glorot et~al.(2011)Glorot, Bordes, and Bengio]{glorot2011deep}
X.~Glorot, A.~Bordes, and Y.~Bengio, ``Deep sparse rectifier neural networks,''
  in \emph{Proceedings of the International Conference on Artificial
  Intelligence and Statistics}, 2011, pp. 315--323.

\bibitem[Srivastava et~al.(2014)Srivastava, Hinton, Krizhevsky, Sutskever, and
  Salakhutdinov]{srivastava2014dropout}
N.~Srivastava, G.~E. Hinton, A.~Krizhevsky, I.~Sutskever, and R.~Salakhutdinov,
  ``Dropout: a simple way to prevent neural networks from overfitting.''
  \emph{Journal of Machine Learning Research}, vol.~15, no.~1, pp. 1929--1958,
  2014.

\bibitem[Krizhevsky et~al.(2012)Krizhevsky, Sutskever, and
  Hinton]{krizhevsky2012imagenet}
A.~Krizhevsky, I.~Sutskever, and G.~E. Hinton, ``{ImageNet} classification with
  deep convolutional neural networks,'' in \emph{Advances in Neural Information
  Processing Systems}, 2012, pp. 1097--1105.

\bibitem[Dahl et~al.(2013)Dahl, Sainath, and Hinton]{dahl2013improving}
G.~E. Dahl, T.~N. Sainath, and G.~E. Hinton, ``Improving deep neural networks
  for {LVCSR} using rectified linear units and dropout,'' in \emph{Proceedings
  of the IEEE International Conference on Acoustics, Speech and Signal
  Processing}, 2013, pp. 8609--8613.

\bibitem[Poole et~al.(2016)Poole, Lahiri, Raghu, Sohl-Dickstein, and
  Ganguli]{poole2016exponential}
B.~Poole, S.~Lahiri, M.~Raghu, J.~Sohl-Dickstein, and S.~Ganguli, ``Exponential
  expressivity in deep neural networks through transient chaos,'' in
  \emph{Advances in Neural Information Processing Systems}, 2016, pp.
  3360--3368.

\bibitem[Schoenholz et~al.(2017)Schoenholz, Gilmer, Ganguli, and
  Sohl-Dickstein]{schoenholz2016deep}
S.~S. Schoenholz, J.~Gilmer, S.~Ganguli, and J.~Sohl-Dickstein, ``Deep
  information propagation,'' \emph{Proceedings of the International Conference
  on Learning Representations}, 2017.

\bibitem[Yang and Schoenholz(2017)]{yang2017mean}
G.~Yang and S.~Schoenholz, ``Mean field residual networks: On the edge of
  chaos,'' in \emph{Advances in Neural Information Processing Systems}, 2017,
  pp. 7103--7114.

\bibitem[Xiao et~al.(2018)Xiao, Bahri, Sohl-Dickstein, Schoenholz, and
  Pennington]{xiao2018dynamical}
L.~Xiao, Y.~Bahri, J.~Sohl-Dickstein, S.~S. Schoenholz, and J.~Pennington,
  ``Dynamical isometry and a mean field theory of \uppercase{CNN}s: How to
  train 10,000-layer vanilla convolutional neural networks,'' \emph{Proceedings
  of the International Conference on Machine Learning}, 2018.

\bibitem[Chen et~al.(2018)Chen, Pennington, and Schoenholz]{chen2018dynamical}
M.~Chen, J.~Pennington, and S.~S. Schoenholz, ``Dynamical isometry and a mean
  field theory of \uppercase{RNN}s: Gating enables signal propagation in
  recurrent neural networks,'' \emph{Proceedings of the International
  Conference on Machine Learning}, 2018.

\bibitem[Hayou et~al.(2018)Hayou, Doucet, and Rousseau]{hayou2018selection}
S.~Hayou, A.~Doucet, and J.~Rousseau, ``On the selection of initialization and
  activation function for deep neural networks,'' \emph{arXiv preprint
  arXiv:1805.08266}, 2018.

\bibitem[Simonyan and Zisserman(2014)]{simonyan2014very}
K.~Simonyan and A.~Zisserman, ``Very deep convolutional networks for
  large-scale image recognition,'' \emph{arXiv preprint arXiv:1409.1556}, 2014.

\end{thebibliography}

\newpage

\appendix
\section*{Supplementary Material}
\addcontentsline{toc}{section}{Appendices}
\renewcommand{\thesubsection}{\Alph{subsection}}

In this section, we provide additional details of derivations and experimental results presented in the paper.

\subsection{Signal propagation in noise regularised neural networks}

 To review, given an input $\mathbf{x}^0 \in \mathbb{R}^{D_0}$, we consider the following noisy random network model
\begin{align}
	\tilde{\mathbf{h}}^l = W^l(\mathbf{x}^{l-1}\odot \epsilon^{l-1}) + \mathbf{b}^l, \textcolor{white}{spa} \text{ for } l = 1, ..., L
	\label{eq: noisy deep model supmat}
\end{align}
where we inject noise into the model using the operator~$\odot$ to denote either addition or multiplication. The vector $\epsilon^l$ is an input noise vector, sampled from a pre-specified noise distribution.
For additive noise, the distribution is assumed to be zero mean. Whereas for multiplicative noise distributions, the mean is assumed to be equal to one. 
The weights $W^l \in \mathbb{R}^{D_l \times D_{l-1}}$ and biases $\mathbf{b}^l \in \mathbb{R}^{D_l}$ are sampled i.i.d.\ from zero mean Gaussian distributions with variances $\sigma^2_w/D_{l-1}$ and $\sigma^2_b$, respectively, where $D_l$ denotes the dimensionality of the $l^{th}$ hidden layer in the network.
The hidden layer activations $\mathbf{x}^l = \phi(\tilde{\mathbf{h}}^l)$ are computed element-wise using an activation function $\phi(\cdot)$, for layers $l = 1, ..., L$.

\subsubsection{Single input signal propagation} \label{sec: single input noisy signal prop}

We consider the network's behavior at initialisation. 
In this setting, the expected mean (over the weights, biases and noise distribution) of a unit in the pre-activations $\tilde{\mathbf{h}}^l_j$ for a single signal passing through the network will be zero with variance
\begin{align*}
  \tilde{q}^l & = \mathbb{E}_{\mathbf{w}, \mathbf{b}, \epsilon}[(\tilde{\mathbf{h}}^l_j)^2] \\ 
  & = \mathbb{E}_{\mathbf{w}, \epsilon}[\{\mathbf{w}^{l,j}\cdot (\mathbf{x}^{l-1}_j \odot \epsilon^{l-1}_j)\}^2] + \mathbb{E}_{\mathbf{b}}[(\mathbf{b}^l_j)^2] \\
	& = \sigma^2_{w}\frac{1}{D_{l-1}}\sum^{D_{l-1}}_{j=1} \left [ \phi(\tilde{\mathbf{h}}^{l-1}_{j})^2 \odot \mathbb{E}_\epsilon[(\epsilon^{l-1}_j)^2] \right ] + \sigma^2_b \enspace ,
\end{align*}
where we use $\mathbf{w}^{l,j}$ to denote the $j$-th row of $W^l$.
The second last line relies on the bias distribution being zero mean, while the final step makes use of the independence between the inputs and the noise in the multiplicative case, and the noise being zero mean in the additive case.
Furthermore, to ensure the expected value of the pre-activations remain unbiased, we only consider additive noise distributions with zero mean and multiplicative noise distributions with a mean equal to one. As in \citet{poole2016exponential}, we make the self averaging assumption and consider the large layer width case where the previous layer's pre-activations are assumed to be Gaussian with zero mean and variance $\tilde{q}^{l-1}$. This gives the following noisy variance map
\begin{align}
	\tilde{q}^l & = \sigma^2_w \left \{ \mathbb{E}_z \left [ \phi\left(\sqrt{\tilde{q}^{l-1}}z \right)^2 \right ] \odot \mu^{l-1}_2 \right \} + \sigma^2_b, 
	\label{eq: noisy variance relation supmat}
\end{align}
where $z \sim \mathcal{N}(0,1)$ and $\mu^l_2 = \mathbb{E}_\epsilon[(\epsilon^{l})^2]$ is the second moment of the noise distribution being sampled from at layer $l$. The initial input variance is given by $q^0 = \frac{1}{D_0}\mathbf{x}^0\cdot\mathbf{x}^0$. 

\subsubsection{Two input signal propagation} \label{sec: two input noisy signal prop}

To study the behaviour of a pair of signals, $\mathbf{x}^{0,a}$ and $\mathbf{x}^{0,b}$, passing through the network, we can compute the covariance in expectation over the noise and the parameters as
\begin{align*}
  \tilde{q}_{ab}^{l} & = \mathbb{E}_{\mathbf{w}, \mathbf{b}, \epsilon}[\tilde{\mathbf{h}}^{l,a}_j\tilde{\mathbf{h}}^{l,b}_j] \\
  & = \mathbb{E}_{\mathbf{w}, \mathbf{b}, \epsilon}\left [\left(\mathbf{w}^{l,j}\cdot (\mathbf{x}^{l-1,a}_j \odot \epsilon^{l-1,a}_j) + \mathbf{b}^{l}_j\right) \left(\mathbf{w}^{l,j}\cdot (\mathbf{x}^{l-1,b}_j \odot \epsilon^{l-1,b}_j) + \mathbf{b}^{l}_j\right) \right] \\
  & = \mathbb{E}_{\mathbf{w}, \mathbf{b}, \epsilon}\left [\left(\mathbf{w}^{l,j}\cdot (\mathbf{x}^{l-1,a}_j \odot \epsilon^{l-1,a}_j)\right) \left(\mathbf{w}^{l,j}\cdot (\mathbf{x}^{l-1,b}_j \odot \epsilon^{l-1,b}_j)\right) \right] \\
  & \textcolor{white}{white} + \mathbb{E}_{\mathbf{w}, \mathbf{b}, \epsilon}\left [\left(\mathbf{w}^{l,j}\cdot (\mathbf{x}^{l-1,a}_j \odot \epsilon^{l-1,a}_j)\right)\mathbf{b}^{l}_j\right] \\
  & \textcolor{white}{white} + \mathbb{E}_{\mathbf{w}, \mathbf{b}, \epsilon}\left [\left(\mathbf{w}^{l,j}\cdot (\mathbf{x}^{l-1,b}_j \odot \epsilon^{l-1,b}_j)\right)\mathbf{b}^{l}_j\right] \\
  & \textcolor{white}{white} + \mathbb{E}_{\mathbf{w}, \mathbf{b}, \epsilon}\left [(\mathbf{b}^{l}_j)^2\right].
\end{align*}
Since the noise is i.i.d and we have that $\mathbb{E}_{\mathbf{b}}[\mathbf{b}^l_j] = 0$, we find that
\begin{align}
  \tilde{q}_{ab}^{l} & = \mathbb{E}_{\mathbf{w}}\left [\left(\mathbf{w}^{l,j}\cdot \mathbf{x}^{l-1,a}_j\right) \left(\mathbf{w}^{l,j}\cdot \mathbf{x}^{l-1,b}_j\right) \right] + \mathbb{E}_{\mathbf{b}}\left [(\mathbf{b}^{l}_j)^2\right]\\
  & = \sigma^2_{w}\frac{1}{D_{l-1}}\sum^{D_{l-1}}_{j=1} \left [ \phi\left(\tilde{\mathbf{h}}^{l-1,a}_j\right)\phi\left(\tilde{\mathbf{h}}^{l-1,b}_j\right) \right] + \sigma^2_b,
\end{align}
which in the large width limit becomes
\begin{align}
	\tilde{q}^l_{ab} = \sigma^2_w  \mathbb{E}_{z_1} \left [ \mathbb{E}_{z_2} \left [ \phi(\tilde{u}_1) \phi(\tilde{u}_2) \right ] \right ] + \sigma^2_b
	\label{eq: corr map supmat}
\end{align}
where $\tilde{u}_1 = \sqrt{\tilde{q}^{l-1}_{aa}}z_1$ and $\tilde{u}_2 = \sqrt{\tilde{q}^{l-1}_{bb}}\left [ \tilde{c}^{l-1}z_1 + \sqrt{1-(\tilde{c}^{l-1})^2}z_2 \right ]$, with the correlation between inputs at layer $l$ given by 
\begin{align}
	\tilde{c}^l = \tilde{q}^l_{ab}/\sqrt{\tilde{q}^l_{aa}\tilde{q}^l_{bb}}.
\end{align}
Here, $z_i \sim \mathcal{N}(0, 1)$ for $i = 1,2$ and $q^l_{aa}$ is the variance of $\tilde{\mathbf{h}}^{l,a}_j$.

\subsection{Signal propagation in noise regularised ReLU networks}

In this section, we give additional details of theoretical results presented in the paper that were specifically derived for noisy ReLU networks.

\subsubsection{Variance of input signals} \label{sec: var dynamics for noisy ReLU networks}

Let $f(z) = \frac{e^{-z^2/2}}{\sqrt{2\pi}}$, then the variance map in \eqref{eq: noisy variance relation supmat} using ReLU, $i.e.$ $\phi(a) = \text{max}(0, a)$, becomes
\begin{align}
  \tilde{q}^l & = \sigma^2_w \left [\int^\infty_{-\infty} f(z) \phi\left(\sqrt{\tilde{q}^{l-1}}z \right)^2 dz \right ] \odot \mu_2 + \sigma^2_b \nonumber \\
  & = \sigma^2_w \left [\int^0_{-\infty} f(z) \phi\left(\sqrt{\tilde{q}^{l-1}}z \right)^2 dz + \int^{\infty}_0 f(z) \phi\left(\sqrt{\tilde{q}^{l-1}}z \right)^2 dz \right ] \odot \mu_2 + \sigma^2_b \nonumber \\
	& = \sigma^2_w \left [\tilde{q}^{l-1} \int^{\infty}_0 f(z) z^2 dz  \right ] \odot \mu_2 + \sigma^2_b \nonumber \\
	& = \sigma^2_w\left [\frac{\tilde{q}^{l-1}}{2} \odot \mu_2 \right ] + \sigma^2_b. 
	\label{eq: var map relu supmat}
\end{align}

\subsubsection{Correlation between input signals} \label{sec: correlation dynamics for noisy ReLU networks}

Assuming that the variance map in \eqref{eq: var map relu supmat} is at its fixed point $\tilde{q}^*$, which exits only if $\sigma^2_w = \frac{2}{\mu_2}$, the correlation map in \eqref{eq: corr map supmat} for a noisy ReLU network is given by
\begin{align}
	\tilde{c}^l = \frac{2}{\mu_2 \tilde{q}^*} \int^\infty_{-\infty} \int^\infty_{-\infty}f(z_1) f(z_2) \phi(\tilde{u}_1) \phi(\tilde{u}_2)dz_2dz_1 + \sigma^2_b
	\label{eq: relu corr map}
\end{align}
where $\phi(a) = \text{max}(a, 0)$, $f(z_i) = \frac{e^{-z_i^2/2}}{\sqrt{2\pi}}$, $\tilde{u}_1 = \sqrt{\tilde{q}^*}z_1$ and $\tilde{u}_2 = \sqrt{\tilde{q}^*}\left[ \tilde{c}^{l-1}z_1 + \sqrt{1-(\tilde{c}^{l-1})^2}z_2 \right ]$. Note that
\begin{align*}
		\tilde{u}_1 & \begin{cases} \geq 0, \text{if } z_1 > 0 \\
		< 0, \text{Otherwise}
			\end{cases} \\
		\tilde{u}_2 & \begin{cases} \geq 0, \text{if } z_2 > \frac{-\tilde{c}^{l-1}z_1}{\sqrt{1-(\tilde{c}^{l-1})^2}} \\
		< 0, \text{Otherwise}
			\end{cases},
\end{align*}
therefore \eqref{eq: relu corr map} becomes
\begin{align}
	\tilde{c}^l & = \frac{2}{\mu_2 \tilde{q}^*} \int^\infty_{0} \int^\infty_{\frac{-\tilde{c}^{l-1}z_1}{\sqrt{1-(\tilde{c}^{l-1})^2}}} f(z_1) f(z_2) \tilde{u}_1 \tilde{u}_2dz_2dz_1  + \sigma^2_b \nonumber \\
	& = \frac{2}{\mu_2 \tilde{q}^*}\sigma^2_w \int^\infty_{0} \int^\infty_{\frac{-\tilde{c}^{l-1}z_1}{\sqrt{1-(\tilde{c}^{l-1})^2}}} f(z_1) f(z_2) \sqrt{\tilde{q}^*}z_1 \sqrt{\tilde{q}^*}\left[ \tilde{c}^{l-1}z_1 + \sqrt{1-(\tilde{c}^{l-1})^2}z_2  \right ]dz_2dz_1 + \sigma^2_b \nonumber \\
	& = \frac{2\tilde{c}^{l-1}}{\mu_2} \int^\infty_{0} \int^\infty_{\frac{-\tilde{c}^{l-1}z_1}{\sqrt{1-(\tilde{c}^{l-1})^2}}} f(z_1) f(z_2) z^2_1 dz_2dz_1 \nonumber \\ 
	& \textcolor{white}{add some white space here} + \frac{2\sqrt{1-(\tilde{c}^{l-1})^2}}{\mu_2} \int^\infty_{0} \int^\infty_{\frac{-\tilde{c}^{l-1}z_1}{\sqrt{1-(\tilde{c}^{l-1})^2}}} f(z_1) f(z_2) z_1z_2 dz_2dz_1.
	\label{eq: corr map linear}
\end{align}
The first term in \eqref{eq: corr map linear} can then be written as
\begin{align}
	\frac{2\tilde{c}^{l-1}}{\mu_2} \left \{ \int^\infty_{0} \int^0_{\frac{-\tilde{c}^{l-1}z_1}{\sqrt{1-(\tilde{c}^{l-1})^2}}} f(z_1) f(z_2) z^2_1dz_2dz_1  + \int^\infty_{0} \int^\infty_0 f(z_1) f(z_2) z^2_1dz_2dz_1 \right \}.
	\label{eq: inter 1}
\end{align}
In \eqref{eq: inter 1}, the first term inside the braces is given by
\begin{align}
	\int^\infty_{0} \int^0_{\frac{-\tilde{c}^{l-1}z_1}{\sqrt{1-(\tilde{c}^{l-1})^2}}} f(z_1) f(z_2) z^2_1dz_2dz_1 & = \frac{1}{2} \int^\infty_{0} f(z_1) z^2_1 \text{erf}\left (\frac{\tilde{c}^{l-1}z_1}{\sqrt{1-(\tilde{c}^{l-1})}} \right)dz_1 \nonumber \\
	& = \frac{1}{2\pi}\left [ \tilde{c}^{l-1}\sqrt{1 - (\tilde{c}^{l-1})^2} + \text{tan}^{-1} \left ( \frac{\tilde{c}^{l-1}}{\sqrt{1 - (\tilde{c}^{l-1})^2}} \right ) \right ] \nonumber \\
	& = \frac{1}{2\pi}\left [ \tilde{c}^{l-1}\sqrt{1 - (\tilde{c}^{l-1})^2} + \text{sin}^{-1} \left ( \tilde{c}^{l-1} \right ) \right ]
	\label{eq: corr map first term}
\end{align}
with $\text{erf}(a) = \frac{1}{\pi}\int^a_{-a} e^{-t^2} dt$. The second term inside the braces in \eqref{eq: inter 1} equals
\begin{align}
	\int^\infty_{0} \int^\infty_0 f(z_1) f(z_2) z^2_1 dz_2dz_1 & = \frac{1}{2} \int^\infty_{0} f(z_1) z^2_1 dz_1\nonumber \\
		& = \frac{1}{4}.
		\label{eq: corr map second term}
\end{align}
Therfore, \eqref{eq: inter 1} becomes
\begin{align}
	\frac{(\tilde{c}^{l-1})^2}{\mu_2 \pi}\sqrt{1 - (\tilde{c}^{l-1})^2} + \frac{\tilde{c}^{l-1}}{\mu_2 \pi} \text{sin}^{-1} \left ( \tilde{c}^{l-1} \right )
	+ \frac{\tilde{c}^{l-1}}{2\mu_2}
	\label{eq: inter 1 final}
\end{align}

Similarly, the second term in \eqref{eq: corr map linear} can be split up as follows
\begin{align}
	\frac{2\sqrt{1-(\tilde{c}^{l-1})^2}}{\mu_2} \left \{ \int^\infty_{0} \int^0_{\frac{-\tilde{c}^{l-1}z_1}{\sqrt{1-(\tilde{c}^{l-1})^2}}} f(z_1) f(z_2) z_1z_2 dz_2dz_1 +
	\int^\infty_{0} \int^\infty_0 f(z_1) f(z_2) z_1z_2 dz_2dz_1 \right \}.
	\label{eq: inter 2}
\end{align}
The first term inside the braces of \eqref{eq: inter 2} is
\begin{align}
	\int^\infty_{0} \int^0_{\frac{-\tilde{c}^{l-1}z_1}{\sqrt{1-(\tilde{c}^{l-1})^2}}} f(z_1) f(z_2) z_1z_2 dz_2dz_1 & =
	\frac{1}{\sqrt{2\pi}} \int^\infty_{0} f(z_1) z_1 \left [ e^{-\frac{\tilde{c}^{l-1}z^2_1}{2\left (1-(\tilde{c}^{l-1})^2 \right)}} - 1 \right] dz_1 \nonumber \\
	& = \frac{1}{\sqrt{2\pi}} \left \{ \frac{1-(\tilde{c}^{l-1})^2}{\sqrt{2\pi}} - \frac{1}{\sqrt{2\pi}}  \right \} \nonumber \\
	& = -\frac{(\tilde{c}^{l-1})^2}{2\pi}
\end{align}
and the second term is
\begin{align}
	\int^\infty_{0} \int^\infty_0 f(z_1) f(z_2) z_1z_2 dz_2dz_1 & = \frac{1}{\sqrt{2\pi}}\int^\infty_{0} f(z_1) z_1 dz_1 \nonumber \\
	& = \frac{1}{2\pi}.
\end{align}
Putting these two terms together, \eqref{eq: inter 2} becomes
\begin{align}
	-\frac{(\tilde{c}^{l-1})^2}{\mu_2\pi}\sqrt{1-(\tilde{c}^{l-1})^2} + \frac{1}{\mu_2\pi}\sqrt{1-(\tilde{c}^{l-1})^2}.
	\label{eq: inter 2 final}
\end{align}
Finally, summing all the terms in \eqref{eq: inter 1 final} and \eqref{eq: inter 2 final} gives \eqref{eq: relu corr map} as
\begin{align}
	\tilde{c}^l = \frac{1}{\mu_2} \left \{ \frac{\tilde{c}^{l-1}\text{sin}^{-1} \left ( \tilde{c}^{l-1} \right ) + \sqrt{1-(\tilde{c}^{l-1})^2}}{\pi}
   + \frac{\tilde{c}^{l-1}}{2} \right \}.
    \label{eq: relu corr map supmat}
\end{align}
We note that for the noiseless case, \eqref{eq: relu corr map supmat} is identical to the result recently obtained by \cite{hayou2018selection}, where the authors used a slightly different approach.

\subsubsection{Depth scales for trainability} \label{sec: Limiting trainability depth scales}

We recap the result in \cite{schoenholz2016deep} and adapt the derivation for the specific case of a noisy ReLU network. Let $c^l = c^* + \varepsilon^l$, such that as long as $\text{lim}_{l \rightarrow \infty}c^l = c^*$ exist we have that $\varepsilon \rightarrow 0$ as $l \rightarrow \infty$. Then \cite{schoenholz2016deep} derived the following asymptotic recurrence relation
\begin{align}
	\varepsilon^{l+1} = \varepsilon^l\chi(c^*) + \mathcal{O}((\varepsilon^l)^2),
	\label{eq: depthscale recurrence}
\end{align}
where 
\begin{align}
	\chi(c^*) = \sigma^2_w \mathbb{E}_{z_1} \left [ \mathbb{E}_{z_2} \left [ \phi^{\prime}(\tilde{u}^*_1)\phi^{\prime}(\tilde{u}^*_2) \right ] \right], 
\end{align}
with $\tilde{u}^*_1 = \tilde{u}_1 = \sqrt{\tilde{q}^*}z_1$ and $\tilde{u}^*_2 = \sqrt{\tilde{q}^*}\left[ \tilde{c}^{*}z_1 + \sqrt{1-(\tilde{c}^{*})^2}z_2 \right ]$. Now, specifically for a noisy ReLU network where $\sigma^2_w = \frac{2}{\mu_2}$, we have that

\begin{align}
	\chi(c^*) & = \frac{2}{\mu_2} \int^\infty_{-\infty} \int^\infty_{-\infty}f(z_1) f(z_2) \phi^{\prime}(\tilde{u}^*_1)\phi^{\prime}(\tilde{u}^*_2)dz_2dz_1 \nonumber \\
	& = \frac{2}{\mu_2} \int^\infty_0 \int^\infty_{-\frac{c^* z_1}{\sqrt{1-(c^*)^2}}}f(z_1) f(z_2) dz_2dz_1 \nonumber \\
	& = \frac{2}{\mu_2} \int^\infty_0 f(z_1) \frac{1}{2} \left [ \text{erf}\left( \frac{c^*z_1}{\sqrt{2}\sqrt{1-(c^*)^2}} \right) + 1 \right ]dz_1 \nonumber \\
	& =  \frac{2}{\mu_2} \left [ \frac{1}{2\pi}\text{tan}^{-1}\left( \frac{c^*}{\sqrt{1-(c^*)^2}} \right) + \frac{1}{4} \right ] \nonumber \\
  & =  \frac{1}{\mu_2\pi} \left [ \text{sin}^{-1}\left( c^* \right) + \frac{\pi}{2} \right ]
  \label{eq: chi supmat}
\end{align}

Note that $\chi(c^*)$ is a constant, thus for large $l$ the solution to the recurrence relation in \eqref{eq: depthscale recurrence} is expected to be exponential, \textit{i.e.} $\varepsilon^l \sim e^{-l/\xi_c}$. Here $\xi_c$, is considered the \textit{depth scale}, which controls how deep discriminatory information about the inputs can propagate through the network. We can then solve for $\xi_c$ to find

\begin{align}
	\xi_c = -1/\text{ln}(\chi(c^*)) = - \text{ln} \left [ \frac{\text{sin}^{-1}\left( c^* \right)}{\mu_2\pi} + \frac{1}{2\mu_2} \right]^{-1}.
\end{align}

\subsubsection{Variance of error signals} \label{sec: backpropagation dynamics for a single input}

Under the mean field assumption, \cite{schoenholz2016deep} approximates the error signal at layer $l$ by a zero mean Gaussian with variance 
\begin{align}
	\tilde{q}^l_\delta = \tilde{q}^{l+1}_\delta\frac{D_{l+1}}{D_l}\sigma^2_w \mathbb{E}_z \left [ \phi^{\prime}\left(\sqrt{\tilde{q}^l}z\right)^2 \right ],
\end{align}
where $\tilde{q}^l_\delta = \mathbb{E}[(\tilde{\delta}^l_i)^2]$, with $\tilde{\delta}^l_i = \phi^\prime(\tilde{\mathbf{h}}^l_i)\sum^{D_{l+1}}_{j=1}\tilde{\delta}^{l+1}_jW^{l+1}_{ji}$. In our context, for a critically initialised noisy ReLU network we have that
\begin{align}
  \tilde{q}^l_\delta & = \tilde{q}^{l+1}_\delta\frac{D_{l+1}}{D_l} \frac{2}{\mu_2}\int^{\infty}_0 f(z) dz \\
  & = \tilde{q}^{l+1}_\delta\frac{D_{l+1}}{D_l} \frac{1}{\mu_2} .
\end{align}

\subsubsection{Correlation between error signals} \label{sec: backpropagation dynamics for a pair of inputs}

The covariance between error signals is approximated using
\begin{align}
	\tilde{q}^l_{ab, \delta} = \tilde{q}^{l+1}_{ab, \delta} \frac{D_{l+1}}{D_{l+2}}\sigma^2_w \mathbb{E}_{z_1} \left [ \mathbb{E}_{z_2} \left [ \phi^{\prime}(\tilde{u}_1) \phi^{\prime}(\tilde{u}_2) \right ] \right ],
\end{align}
where $\tilde{u}_1$ and $\tilde{u}_2$ are defined as was done in the forward pass. Here, we simply use the result in \eqref{eq: chi supmat} for noisy ReLU networks to find
\begin{align}
  \tilde{q}^l_{ab, \delta} & = \tilde{q}^{l+1}_{ab, \delta} \frac{D_{l+1}}{D_{l+2}}\chi(c^*) \\
  & = \tilde{q}^{l+1}_{ab, \delta} \frac{D_{l+1} \left [ \text{sin}^{-1}\left( c^* \right) + \frac{\pi}{2} \right ]}{D_{l+2}\mu_2\pi}.
\end{align}

\subsection{Experimental details} \label{sec: experimental details}

In this section we provide additional details regarding our experiments in the paper. Code to reproduce all the experiments is available at \href{https://github.com/ElanVB/noisy\_signal\_prop}{https://github.com/ElanVB/noisy\_signal\_prop}.

\subsubsection{Input data}
For all experiments the network input data properties that remain consistent (unless stated otherwise) are as follows: each observation consists of 1000 features and each feature value is drawn i.i.d. from a standard normal distribution.

\subsubsection{Variance propagation dynamics}
The experiments conducted to gather results for Figures \ref{fig: add noise qmap} and \ref{fig: qmap} aim to empirically show the relationship between the variances at arbitrary layers in a neural network.

\textit{Iterative map}: For the results depicted in Figures \ref{fig: add noise qmap} (a) and \ref{fig: qmap} (a), the experimental set up is as follows.
The data used as input to these experiments comprises of 30 sets of 30 observations.
The input is scaled such that the variance of observations within each set is the same and the variance across each set is different and forms a range of $q_{\text{set}} \in \left [ 0, 15 \right ]$.
As such, our results are averaged over 30 observations and 50 samplings of initial weights to a single hidden-layer network.

\textit{Convergence dynamics}: For the results depicted in Figures \ref{fig: add noise qmap} (b) and \ref{fig: qmap} (b), the experimental set up is as follows.
The data used as input to these experiments comprises of a set of 50 observations scaled such that each observation's variance is four ($q = 4$).
As such, our results are averaged over 50 observations and 50 samplings of initial weights to a 15 hidden-layer network.

\subsubsection{Correlation propagation dynamics}
The experiments conducted to gather results for Figure \ref{fig: cmap} and \ref{fig: theory depth scales} aim to empirically show the relationship between the correlations of observations at arbitrary layers in a neural network.

\textit{Iterative map}: For the results depicted in Figure \ref{fig: cmap} (a), the experimental set up is as follows.
The data used as input to these experiments comprises of 50 sets of 50 observations.
The first observation in each set is sampled from a standard normal distribution and subsequent observations are generated such that the correlation between the first element and the $i^{\text{th}}$ element form a range of $\text{corr}_{0, i} \in [0, 1]$.
As such, our results are averaged over 50 observations and 50 samplings of initial weights to a single hidden-layer network.

\textit{Convergence dynamics}: For the results depicted in Figure \ref{fig: cmap} (b), the experimental set up is as follows.
The data used as input to these experiments comprises of three sets of 50 equally correlated observations.
Each set has a different correlation value such that $\text{corr}_{\text{set}} \in \left \{ 0, 0.5, 0.9 \right \}$.
As such, our results are averaged over 50 observations and 50 samplings of initial weights to a 15 hidden-layer network.

\textit{Confirmation of exponential rate of convergence for correlations}: This section discusses how the results depicted in Figure \ref{fig: theory depth scales} are acquired.
These experiments support the assumption that the rate of convergence for correlations is exponential when using noise regularisation with rectifier neural networks.
The experimental set up for this section is very similar to that of the above convergence dynamics experiment, the only difference being the statistics we calculate from the correlation values.
The aspect of this experiment that may seem the most unclear is the reason why we claim that the negative inverse slope of a linear fit to the $\log$ differences in correlation should match the theoretical values for $\xi_c$.
The derivation to justify this is as follows. If a good fit of the form $al + b$ can be found in the logarithmic domain for the rate of convergence, it would strongly indicate that the convergence rate is exponential.
Following this, we set the problem up like so:

\begin{align*}
  | c^l - c^* | &\approx e^{-l/\xi_c} \\
  \therefore \ln \left ( | c^l - c^* | \right ) &\approx \frac{-l}{\xi_c}.
\end{align*}

Let us now assume that $\ln \left ( | c^l - c^* | \right )$ can be linearly approximated:
\begin{align*}
  \therefore \ln \left ( | c^l - c^* | \right ) &\approx al + b,\\
  \therefore al + b &\approx \frac{-l}{\xi_c},\\
  \therefore \xi_c &\approx \frac{-l}{al + b}.
\end{align*}

Since we are concerned with deep neural networks, we can take the limit as $l$ becomes arbitrarily large and see that as $l$ grows the effect of $b$ decreases $\left ( \lim_{l \to \infty} | al | \gg | b | \right )$.
Thus, we continue like so:
\begin{align*}
  \lim_{l \to \infty} \xi_c &\approx \lim_{l \to \infty} \frac{-l}{al}\\
                            &\approx - \frac{1}{a}.
\end{align*}

Thus, we have come to the finding that if the correlation rate of convergence is exponential and we work with deep neural networks, the negative inverse slope of a linear fit to the $\log$ differences in correlation should match the theoretical values for $\xi_c$.
Figure \ref{fig: theory depth scales} shows that the theory closely matches this approximation.

\subsubsection{Depth scales} \label{sec: var depth scale theory}
This section handles the experiments conducted related to determining the maximum depth variance information can stably propagate through a network and the depth at which these networks can be trained, both depicted in Figure \ref{fig: depth scales}.

The MNIST and CIFAR-10 datasets were used and were pre-processed using standard techniques.
Throughout these experiments mini-batches of $128$ observations were used.

\textit{Variance depth scales}: The experiments depicted in Figures \ref{fig: depth scales} (a) and (d) are interested in testing the numerical stability of networks initialised using different $\sigma_w^2$ values while using $32$-bit floating point numbers.
To test the depth of stable variance propagation, a network with 1000 hidden layers is used.
The network used in this experiment makes use of dropout with $p = 0.6$, where $p$ is the probability of keeping a neuron's value, thus the critical value for $\sigma_w^2$ is $1.2$.
As such, a linearly spaced range of $\sigma_w^2 \in \left [ 0.1, 2.5 \right ]$ is used to select 25 different values.

We use the following approach to predict the depth beyond which variances become numerically unstable. 
At criticality for multiplicative noise $(\sigma_w, \sigma_b) = (\sqrt{2/\mu_2}, 0)$, however, for weights initialised off this critical point \eqref{eq: var map relu supmat} becomes
\begin{align}
  \tilde{q}^l & = \tilde{q}^{l-1} \left ( \frac{\sigma_w^2 \mu_2}{2} \right ) \nonumber \\
  & = \left [ \tilde{q}^{l-2} \left ( \frac{\sigma_w^2 \mu_2}{2} \right ) \right ]\left ( \frac{\sigma_w^2 \mu_2}{2} \right )\nonumber \\
  & = \tilde{q}^0 \left ( \frac{\sigma_w^2 \mu_2}{2} \right )^l.
  \label{eq: stable boundary}
\end{align}
If $\sigma^2_w > \frac{2}{\mu_2}$, we let $\tilde{q}^l = K$, where $K$ is the largest positive number representable by the computer. In our case, using 32-bit floating point precision, this number is equal to $3.4028235\times 10^{38}$. 
Otherwise, if $\sigma^2_w < \frac{2}{\mu_2}$ we select $K = 1.1754944\times 10^{-38}$, the smallest possible positive number. 
Furthermore, let $L^*$ represent the layer $l$ in \eqref{eq: stable boundary} at which the value $K$ is reached, then we can scale our input data such that $\tilde{q}^0 = 1$ and solve for $L^*$ to find
\begin{align}
  L^* = \text{ln}(K)/\text{ln}\left( \frac{\sigma^2_w\mu_2}{2}\right).
\end{align} 
Therefore, we expect numerical instability issues to occur beyond a depth of $L^*$.

\textit{Trainable depth scales}: The experiments depicted in Figures \ref{fig: depth scales} (b), (c), (e) and (f) are concerned with determining at what depth a critically initialised network with a specified dropout rate can train effectively.
To this end, 10 linearly spaced values for dropout on the range $p \in \left [ 0.1, 1.0 \right ]$ and 10 linearly spaced network depths on the integer range $l \in \left [ 2, 40 \right ]$ are tested.

The task presented to the network in this experiment is to learn the identity function within 200 epochs.
As such, the network is set up as an auto-encoder and uses stochastic gradient decent with a learning rate of $10^{-3}$. The input data is divided into a training and validation set, each containing $50000$ and $10000$ observations respectively.

\subsubsection{Additional results} \label{sec: additional results}

In this section we provide some additional experiments on the training dynamics of deep noisy ReLU networks from different initialisations.



\begin{figure}
 \includegraphics[width=\linewidth]{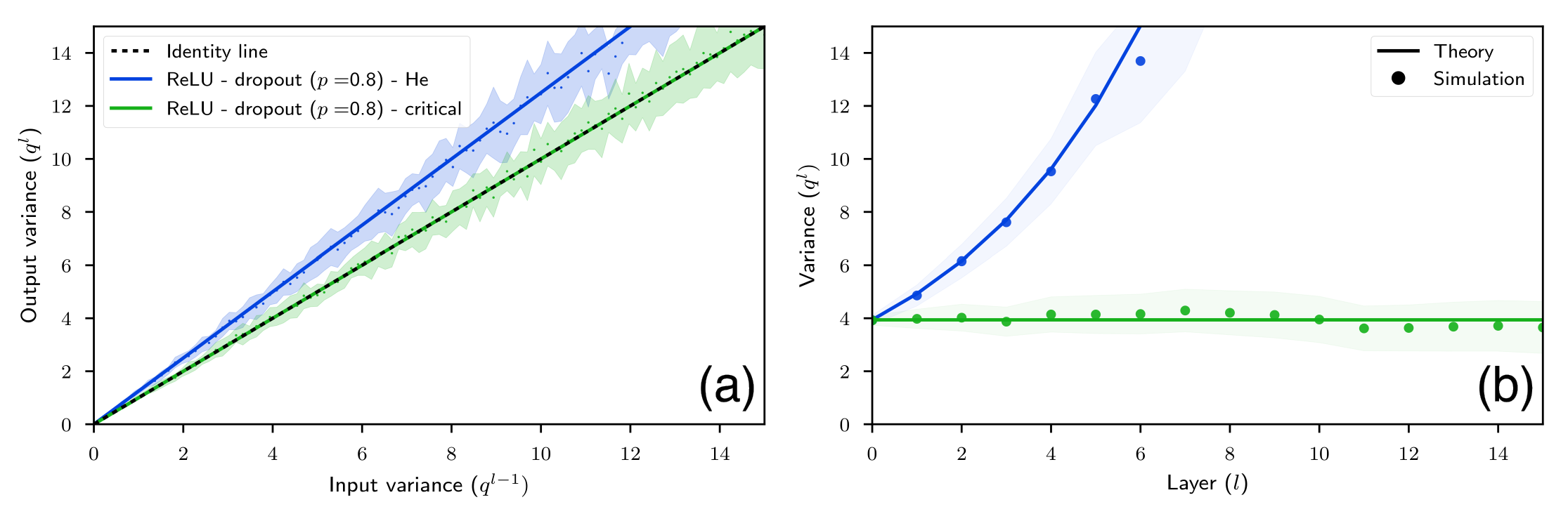}
 \caption{\textit{Critical initialisation for ReLU networks with dropout.} Lines correspond to theoretical predictions and points to numerical simulations, for random ReLU networks with dropout ($p = 0.8$), initialised according to the method proposed by \citet{he2015delving} (blue) and at criticality (green). \textbf{(a)}:~Iterative variance map where the identity line is displayed as a dashed black line. \textbf{(b)}:~Variance dynamics during forward signal propagation.}
 \label{fig: crit inits}
\end{figure}

In Figure \ref{fig: crit inits} we compare the standard ``He'' initialisation (blue) with the critical initialisation (green) for a ReLU network with dropout regularisation $(p = 0.8)$. By not initialising at criticality due to dropout noise, the variance map for the ``He'' strategy no longer lies on the identity line in (a) and as a result, the forward propagating signal can be seen to explode in (b). However, by compensating for the amount of injected noise, the above derived critical initialisation for dropout preserves the signal throughout the entire forward pass, with roughly constant variance dynamics. 

Next, we provide some additional experiments on the trainability of deep ReLU networks with dropout on real-world data sets. 

From our analysis in the paper, we expect that as the depth of the network increases, any initialisation strategy that does not factor in the effects of noise, will cause the forward propagating signal to  become increasingly unstable. 
For very deep networks, this might cause the signal to either explode or vanish, even within the first forward pass, making the network untrainable. 

\begin{figure}
  \centering
  \includegraphics[width=\textwidth]{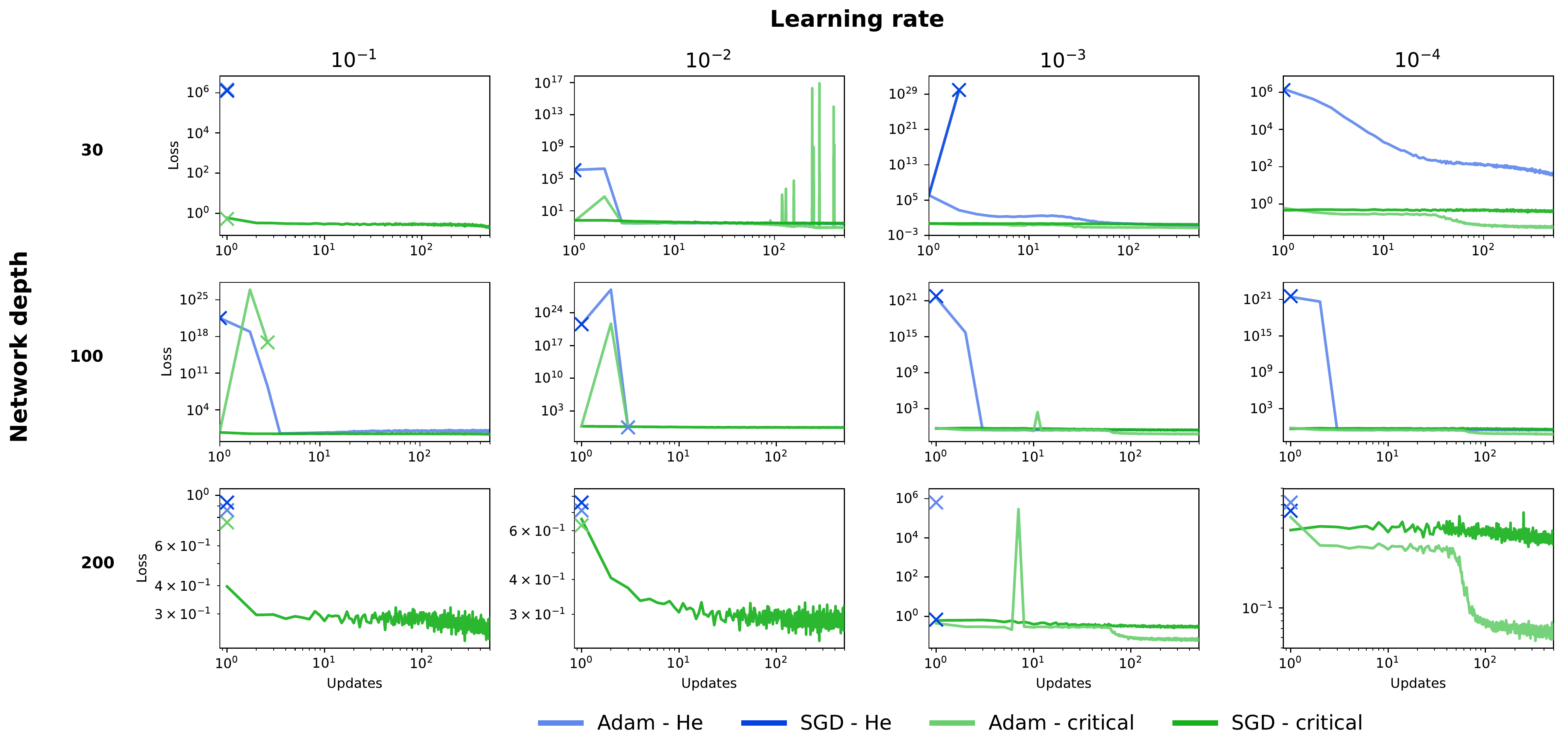}
  \caption{
    Comparing the ``He'' initialisation strategy to critical dropout initialisation for ReLU networks using dropout $(p = 0.6)$ on CIFAR-10.
    While networks initialised at criticality (green) are able to train at large depths $(L = 200)$ as seen in the bottom row, networks initialised with the “He” strategy (blue) become untrainable irrespective of the chosen learning rate or optimisation procedure.
    An ``X'' marks the point at which a network completely stopped training.
    Training losses and number of network updates are shown in log-scale.
  }
  \label{CIFAR10}
\end{figure}

\begin{figure}
  \centering
  \includegraphics[width=\textwidth]{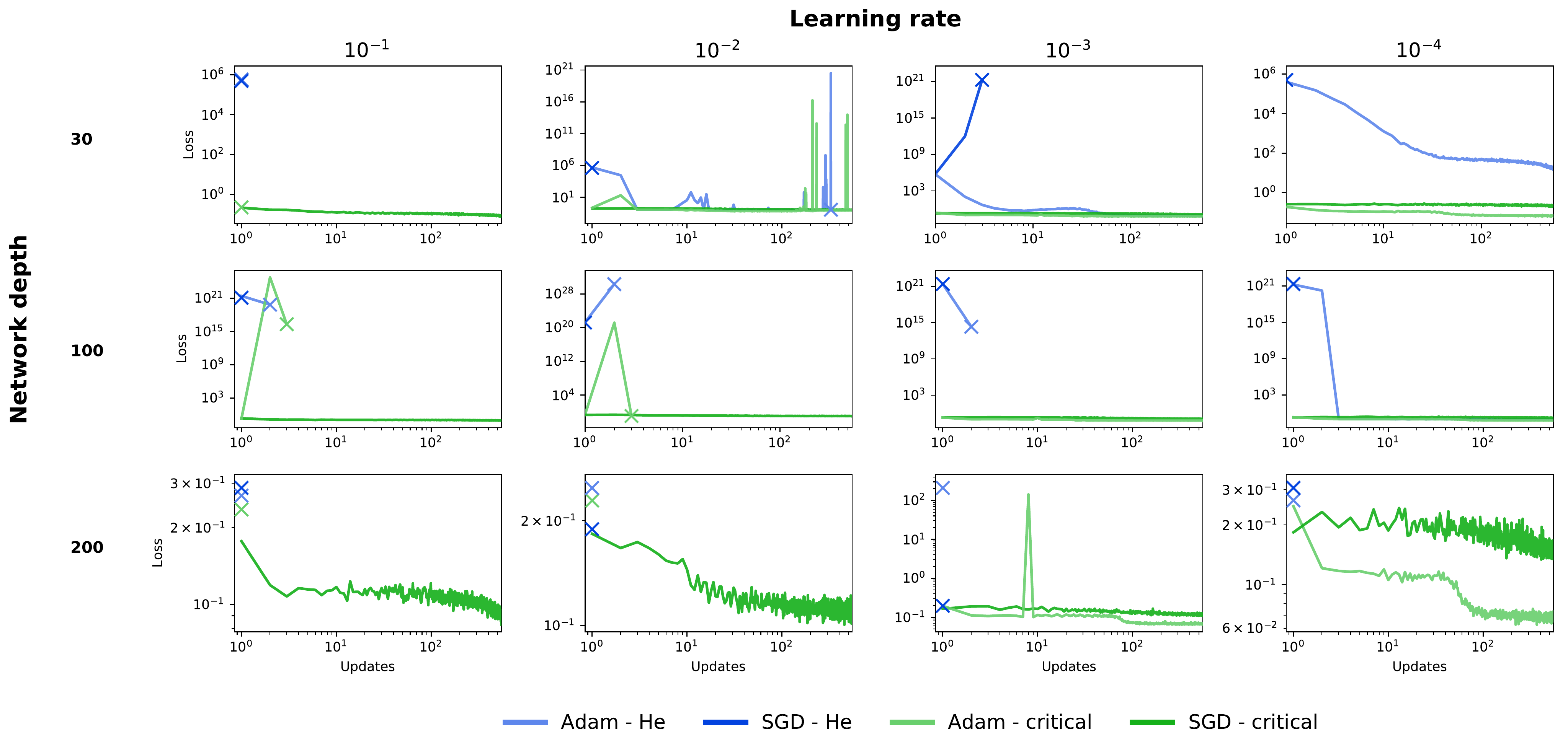}
  \caption{
    Comparing the ``He'' initialisation strategy to critical dropout initialisation for ReLU networks using dropout $(p = 0.6)$ on MNIST.
    While networks initialised at criticality (green) are able to train at large depths $(L = 200)$ as seen in the bottom row, networks initialised with the “He” strategy (blue) become untrainable irrespective of the chosen learning rate or optimisation procedure.
    An ``X'' marks the point at which a network completely stopped training.
    Training losses and number of network updates are shown in log-scale.
  }
  \label{MNIST}
\end{figure}

To test this, we trained a denoising autoencoder network with dropout noise ($p = 0.6$) on MNIST and CIFAR-10 using squared reconstruction loss. We consider several network depths ($L= 30, 100, 200$), learning rates ($\alpha = 0.1, 0.01, 0.001, 0.0001$) and optimisation procedures (SGD and Adam), with $1000$ neurons in each layer. 
The results for training on CIFAR-10 are shown in Figure \ref{CIFAR10} for both the ``He'' intialisation (blue) and the critical dropout initialisation (green).  (For MNIST, see Figure \ref{MNIST}; the core trends and resulting conclusions regarding network trainability is the same for both data sets, which we discuss below.)

As the depth increases, moving from the top to the bottom row in Figure \ref{CIFAR10}, networks initialised at the critical point for dropout seem to remain trainable even up to a depth of 200 layers (we see the loss start to decrease over five epochs). 
In contrast, networks using the ``He'' initialisation become increasingly more difficult to train, with no training taking place at very large depths. 
These findings make sense in terms of the variance dynamics analysed in the paper, however, these experimental successes seem to run counter to our theoretical analysis of trainable depth scales (this contradiction can also be seen in Figure \ref{fig: depth scales}). Understanding this discrepancy is of particular interest to us.

\begin{figure}
	\includegraphics[width=\linewidth]{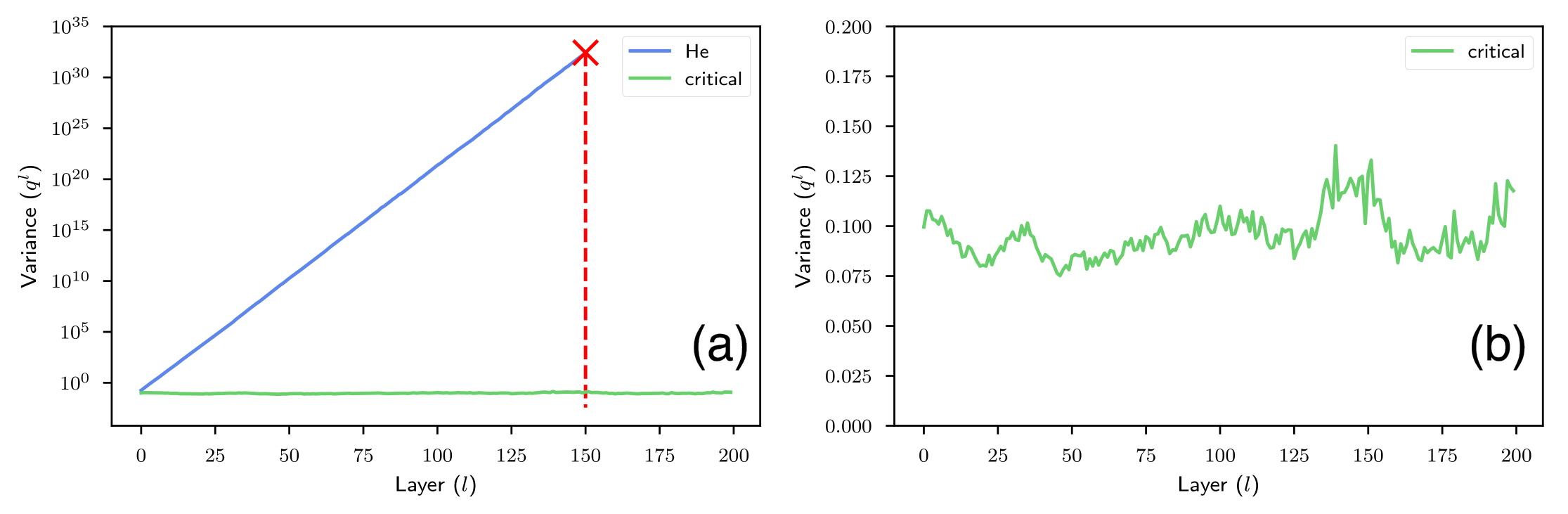}
	\caption{\textit{Variance dynamics for signal propagation in the first forward pass for a 200 layer autoencoder network fed a batch of 500 training examples from CIFAR-10.} \textbf{(a)} Exploding activation variance (blue) reaching overflow levels (marked with a red ``X'') for the ``He'' intialisation, with no signal reaching the output layer (shown in log-scale). \textbf{(b)} Zoomed in display of the roughly constant variance dynamics in (a) for the critical dropout initialisation.}
	\label{fig: cifar10 variance}
\end{figure}

To verify that the lack of training in Figure \ref{CIFAR10} is due to poor signal propagation, we plot the empirical variance of the pre-activations in Figure \ref{fig: cifar10 variance}, for the first forward pass of a 200 layer autoencoder network. For the ``He'' initialisation, the variance in (a) grows rapidly to the point of causing numerical instability and overflow (indicated by the red dashed line), well before any signal is able to reach the output layer. However as shown in (b), by initialising at criticality, signal is able to propagate reliably even at large depths.

\begin{figure}
	\includegraphics[width=\linewidth]{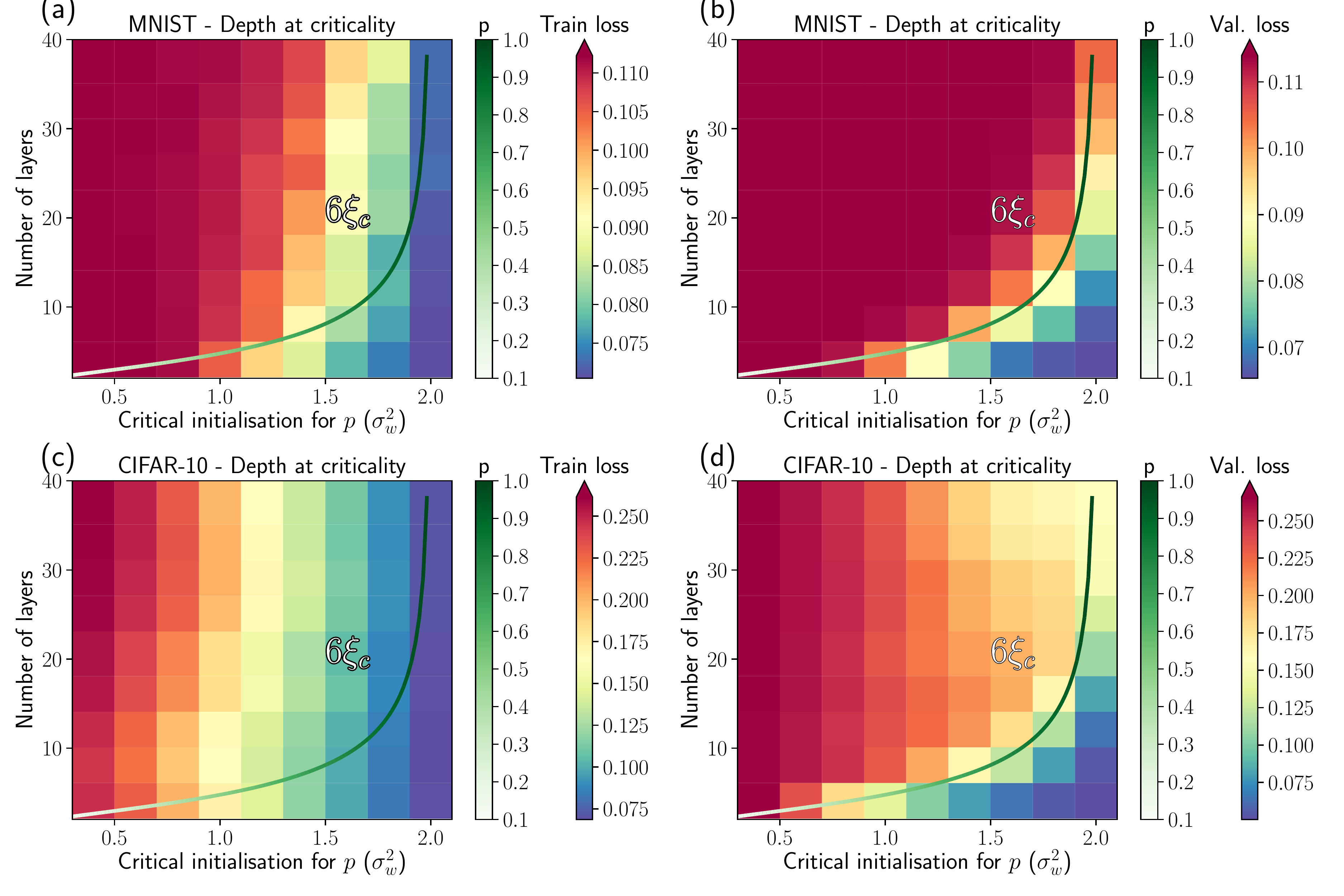}
	\caption{\textit{Depth scale experiments on MNIST and CIFAR-10.} \textbf{(a)} Depth scales fit to the training loss on MNIST for networks initialised at criticality for dropout rates $p=0.1$ (severe dropout) to $p=1$ (no dropout). \textbf{(b)} Depth scales fit to the validation loss on MNIST. \textbf{(c) - (d)}: Similar to (a) - (c), but for CIFAR-10.} 
	\label{fig: depth scales supmat}
\end{figure}

\end{document}